\documentclass{article}

\usepackage{microtype}
\usepackage{graphicx}
\usepackage{subcaption}
\usepackage{booktabs} 
\usepackage{hyperref}

\usepackage[preprint]{icml2026}

\makeatletter
\let\algorithmic\@undefined
\let\endalgorithmic\@undefined
\makeatother

\usepackage{algorithm}
\usepackage{algpseudocode}

\usepackage{amsmath}
\usepackage{amssymb}
\usepackage{mathtools}
\usepackage{amsthm}

\usepackage[capitalize,noabbrev]{cleveref}

\theoremstyle{plain}

\theoremstyle{definition}

\theoremstyle{remark}

\usepackage{graphicx}
\usepackage{hyperref}
\usepackage{url}
\usepackage{amsmath}
\usepackage{cleveref}
\usepackage{multirow}
\usepackage{amsmath,amssymb}
\usepackage[table]{xcolor}
\usepackage{colortbl}
\usepackage{tcolorbox}  
\usepackage{enumitem} 

\definecolor{lightblue}{HTML}{ebf3f8}

\definecolor{mediumblue}{HTML}{d7e8f2}

\definecolor{deepblue}{HTML}{c8dfed}

\definecolor{mplBlue}{RGB}{31, 119, 180}
\definecolor{mplOrange}{RGB}{255, 127, 14}
\definecolor{mplGreen}{RGB}{44, 160, 44}
\usepackage[textsize=tiny]{todonotes}

\icmltitlerunning{From Perception to Action: An Interactive Benchmark for Vision Reasoning}
\definecolor{colorC}{HTML}{0D47A1}  
\definecolor{colorH}{HTML}{1976D2}  
\definecolor{colorA}{HTML}{2196F3}  
\definecolor{colorIN}{HTML}{0097A7} 

\newcommand{\colorCtext}[1]{\textcolor{colorC}{#1}}
\newcommand{\colorHtext}[1]{\textcolor{colorH}{#1}}
\newcommand{\colorAtext}[1]{\textcolor{colorA}{#1}}
\newcommand{\colorINtext}[1]{\textcolor{colorIN}{#1}}

\newcommand{\bench}{%
    \textbf{\textit{%
        \colorCtext{C}\colorHtext{H}\colorAtext{A}\colorINtext{IN}%
}}%
}

\begin{document}

\twocolumn[
  \icmltitle{From Perception to Action: An Interactive Benchmark for Vision Reasoning}



\icmlsetsymbol{equal}{*}
\icmlsetsymbol{lead}{\dagger}

\begin{icmlauthorlist}
  \icmlauthor{Yuhao Wu}{equal,sutd}
  \icmlauthor{Maojia Song}{equal,sutd}
  \icmlauthor{Yihuai Lan}{equal,smu}
  \icmlauthor{Lei Wang}{smu}
  \icmlauthor{Zhiqiang Hu}{sutd}
  \icmlauthor{Yao Xiao}{sutd}
  
  \icmlauthor{Heng Zhou}{ustc}
  \icmlauthor{Weihua Zheng}{sutd}
  \icmlauthor{Dylan Raharja}{sutd}
  \icmlauthor{Soujanya Poria}{ntu}
  \icmlauthor{Roy Ka-Wei Lee}{sutd}
\end{icmlauthorlist}

\icmlaffiliation{sutd}{Singapore University of Technology and Design (SUTD), Singapore}
\icmlaffiliation{smu}{Singapore Management University (SMU), Singapore}
\icmlaffiliation{ntu}{Nanyang Technological University (NTU), Singapore}
\icmlaffiliation{ustc}{University of Science and Technology of China (USTC), China}

\icmlcorrespondingauthor{Lei Wang}{lei.wang.2019@phdcs.smu.edu.sg}
\icmlcorrespondingauthor{Roy Ka-Wei Lee}{roy\_lee@sutd.edu.sg}

\vskip 0.3in
]



\printAffiliationsAndNotice{\icmlEqualContribution}

\begin{abstract}

Understanding the physical structure is essential for real-world applications such as embodied agents, interactive design, and long-horizon manipulation. 
Yet, prevailing Vision–Language Model (VLM) evaluations still center on structure-agnostic, single-turn setups (e.g., VQA), which fail to assess agents' ability to reason about how geometry, contact, and support relations jointly constrain what actions are possible in a dynamic environment.
To address this gap, we introduce the \colorCtext{\textbf{C}}ausal \colorHtext{\textbf{H}}ierarchy of \colorAtext{\textbf{A}}ctions and \colorINtext{\textbf{In}}teractions (\bench) benchmark, an interactive 3D, physics-driven testbed designed to evaluate whether models can understand, plan, and execute structured action sequences grounded in physical constraints. 
\bench~shifts evaluation from passive perception to active problem solving, spanning tasks such as interlocking mechanical puzzles and causal-chain manipulation.
We conduct a comprehensive study of state-of-the-art VLMs and diffusion-based models under unified interactive settings.
Our results show that top-performing models still struggle to internalize physical structure and causal constraints, often failing to produce reliable long-horizon plans and cannot robustly translate perceived structure into effective actions. The project is available at \url{https://social-ai-studio.github.io/CHAIN/}.

\end{abstract}

\section{Introduction}
    \begin{figure}[htb]
    \label{Fig:diff_with_VQA}
    \centering
    \includegraphics[width=0.75\columnwidth]{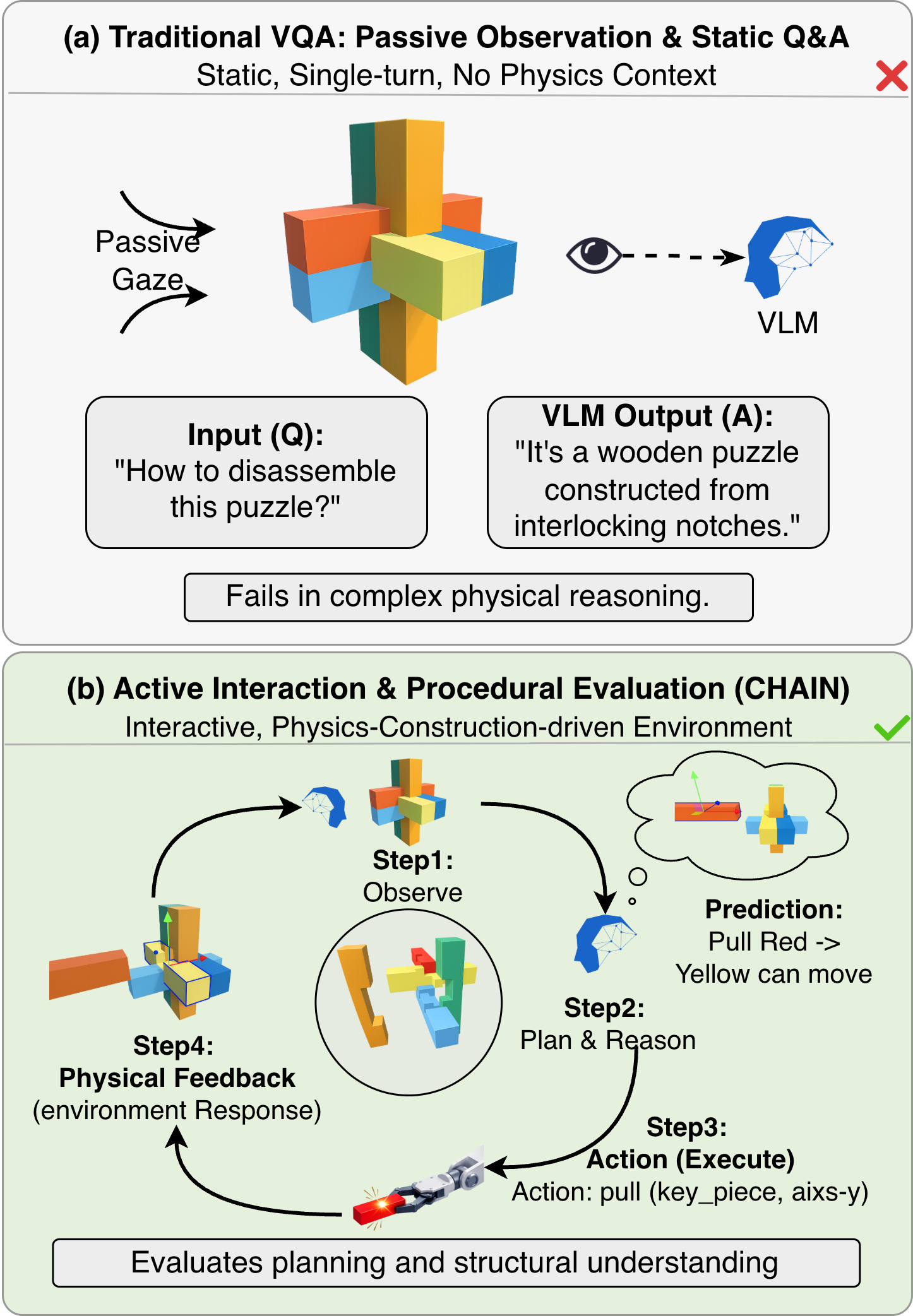}
    
    \caption{
        \textbf{Static vs. Interactive Evaluation of Physical Struction Reasoning.} 
        \textbf{(a)}~Traditional VQA relies on passive observation of an image. 
        \textbf{(b)}~Our paradigm requires multi-step interaction, enabling procedural evaluation of planning and structural understanding.
    }
    \label{Fig:diff_with_VQA}
    
    \vspace{-10pt}
\end{figure}

Real-world physical problem solving takes place in interactive settings involving rich visual feedback and physical constraints, where objects must be manipulated, configurations explored, and outcomes assessed under feasibility constraints.
Success in such settings depends on tightly coupling perception, action, and iterative feedback over multiple steps, rather than on static scene understanding alone~\cite{kim2024openvlaopensourcevisionlanguageactionmodel,bai2025qwen25vltechnicalreport}.
In these scenarios, the central challenge shifts from recognizing what is present to anticipating what actions are feasible, as feasibility is jointly determined by geometric structure, interfacial contacts, and multi-body support relations~\cite{cheng2025embodiedevalevaluatemultimodalllms,li2024behavior1khumancenteredembodiedai,gu2023maniskill2unifiedbenchmarkgeneralizable}.

Given these requirements, an important question is whether existing evaluation protocols adequately capture such interactive, constraint-driven reasoning.
However, current evaluations for Vision--Language Model (VLM) remain largely centered on static, single-turn formats such as Visual Question Answering (VQA), where performance is measured by the correctness of a final textual answer~\citep{agrawal2016vqavisualquestionanswering,lu2022learn,he2024olympiadbench,physbench2025,Chen_2024_CVPR,lyu2024mmscan,wang2023embodiedscan,Guo2024PhyGrasp,xu2411llava}. 
While effective for diagnosing recognition and factual grounding, these benchmarks—by remaining largely static and single-turn—leave untested the agent’s ability to plan and adapt actions as constraints evolve, including whether it can anticipate how early choices will constrain (or preserve) the feasible action space later on; as a result, they tend to underestimate the difficulty of physical reasoning in interactive scenarios~\citep{cheng2025embodiedevalevaluatemultimodalllms}.  In particular, they do not evaluate whether an agent can reason about how early actions constrainor preserve future feasible action spaces.
Alongside VLMs, diffusion-based models have been explored as another paradigm for reasoning and planning.
However, existing evaluations of such approaches predominantly focus on simplified 2D environments that sidestep the challenges introduced by 3D geometry, contact constraints, and support relations~\cite{taufeeque2025planningrecurrentneuralnetwork,yang2025reasoningvideoevaluationvideo,luo2025vreasonbenchunifiedreasoningbenchmark}. 
As a result, it remains unclear whether either modern VLMs or diffusion models can reliably solve structured 3D physical problems whose feasibility is governed by hidden geometric constraints and contact-driven dependencies.

To address this gap, we introduce \bench~(\colorCtext{\textbf{C}}ausal \colorHtext{\textbf{H}}ierarchy of \colorAtext{\textbf{A}}ctions and \colorINtext{\textbf{In}}teractions), an interactive 3D benchmark designed to evaluate whether models can understand, plan, and execute structured action sequences grounded in physical constraints. 
\bench~reorients evaluation from passive perception to active problem solving in a physics--engine-driven environment\footnote{Our physics environment is implemented with Unity~\citep{unity,juliani2020unitygeneralplatformintelligent}.},
where models must iteratively observe the environment, select feasible interactions, and revise their plans based on intermediate outcomes. \bench~comprises task families that stress complementary aspects of structured physical reasoning:
(1) Interlocking mechanical puzzles, which probe constraint-aware reasoning over tightly coupled 3D objects, including contact-rich dependencies, non-intuitive feasible motions, and hidden geometric constraints;
(2) 3D stacking and packing, which assess maintaining global feasibility over long horizons and stability reasoning under gravity, including support relations, balance constraints, and feasible assembly orders for stable configurations.
Cross two families, these tasks emphasize structure-sensitive decision making: correct actions depend on understanding how geometry and constraints restrict what can be done next.

We conduct a comprehensive empirical study of state-of-the-art VLMs and diffusion-based models under a unified interactive setting. 
Our results indicate that current models have difficulty internalizing physical structure and constraint dependencies, often fail to maintain coherent multi-step strategies, and do not robustly translate perceived structure into effective actions over extended interactions. 
Overall, \bench~provides a challenging benchmark and reproducible baselines to catalyze progress toward physically grounded, structure-aware interactive agents.

Our contributions are twofold:
\begin{enumerate}[topsep=1.5pt, itemsep=2pt, parsep=0pt, partopsep=0pt, leftmargin=1.5em]
    \item We introduce and open-source \bench, an interactive 3D, physics-driven benchmark that shifts evaluation from static VQA to closed-loop physical problem solving.
    \bench~comprises 109 distinct interactive levels with clear difficulty separation, designed to test whether agents can infer and exploit physical structure---including hidden geometric constraints, contact-induced dependencies, and multi-body support relations---to select feasible actions over multi-step interaction.

    \item We conduct a unified empirical study of state-of-the-art VLMs and diffusion-based image-to-video models under the same interactive protocol and exposing persistent limitations in physical-structure-aware reasoning.
    In particular, current models often fail to convert perceived structure into valid action sequences as constraints tighten, struggling with feasibility reasoning grounded in 3D geometry, contact constraints, and support relations over long-horizon interactions.
\end{enumerate}

\section{\bench~Benchmark}

This section introduces our interactive 3D benchmark for rigorous evaluation of VLMs' physical reasoning capabilities. We first outline the task families and their targeted capabilities (\Cref{subsec:task}), then present the end-to-end pipeline for building controllable, reproducible environments (\Cref{sub:benchmark-build}), and define metrics for correctness and efficiency under realistic interaction cost constraints (\Cref{subsec:metric})

\subsection{Task Overview} 

\label{subsec:task}

\paragraph{Puzzle (Interlocking Mechanical Structures).}
The puzzle task family drawing inspiration from classical interlocking designs, including \emph{Kongming locks}, \emph{Lu Ban locks}, interlocking cubes, and burr puzzles.
In these tasks, the agent is required to assemble or disassemble multi-piece structures \textbf{mortise-and-tenon}\footnote{Mortise-and-tenon refers to a traditional interlocking joint structure that connects parts without nails or adhesives, widely used in classical Chinese craftsmanship.} through fine-grained manipulation actions (e.g., pick, rotate, insert), where progress critically depends on executing steps in the correct order. 
The key challenge lies in reasoning about mechanical constraints, such as kinematic feasibility, collision avoidance, force directions, and the dependencies between components, making these tasks inherently long-horizon and causally structured.

Our puzzle suite consists of \textbf{32 task instances} organized into three difficulty levels: \textbf{10 easy}, \textbf{12 medium}, and \textbf{10 hard}, with difficulty naturally increasing from small six-piece locks to complex designs involving more than thirty interlocking parts.
Task success is determined by rule-based verification of the final configuration, requiring an exact match to the target state.
Overall, these puzzles closely resemble real-world mechanical assembly and maintenance scenarios, where robust performance demands precise structural understanding, planning, and reliable execution of physically grounded actions.

\paragraph{Stacking (3D Spatial Packing).}
The stacking family focuses on spatial packing problems, where the agent must place multiple blocks with diverse geometric shapes into a fixed container (e.g., a cube or rectangular prism).
At each step, the agent performs fine-grained manipulation operations (e.g., select, rotate, and place blocks) to incrementally construct a valid packing arrangement.
These tasks primarily evaluate three-dimensional spatial reasoning, requiring the agent to jointly reason about shape compatibility, orientation constraints, and how early placement decisions progressively restrict the remaining free space.

Our stacking suite comprises \textbf{77 task instances}, spanning \textbf{10 easy}, \textbf{20 medium}, and \textbf{47 hard} task instances.
The difficulty of the task is jointly determined by the container size, the number of objects and the complexity of the block shape.
Importantly, the stacking environment is generated \textbf{programmatically} (Appendix \ref{app:polycube-gen}), enabling systematic control over container dimensions, block shapes, and object counts.
As a result, stacking instances can be expanded to \textbf{near-unlimited complexity}, with problem difficulty scaling naturally as the geometry becomes more irregular and the feasible solution space becomes more constrained.
Task success is determined by rule-based validation of the final configuration, requiring exact volume coverage without overlap or gaps, making this benchmark a scalable testbed to evaluate long-horizon spatial planning and constraint satisfaction\footnote{We present several examples in Appendix~\ref{app:task_case}.}.

\subsection{Construction Pipeline}
\label{sub:benchmark-build}
\begin{figure*}[t]
  \centering
  \includegraphics[width=0.8\textwidth]{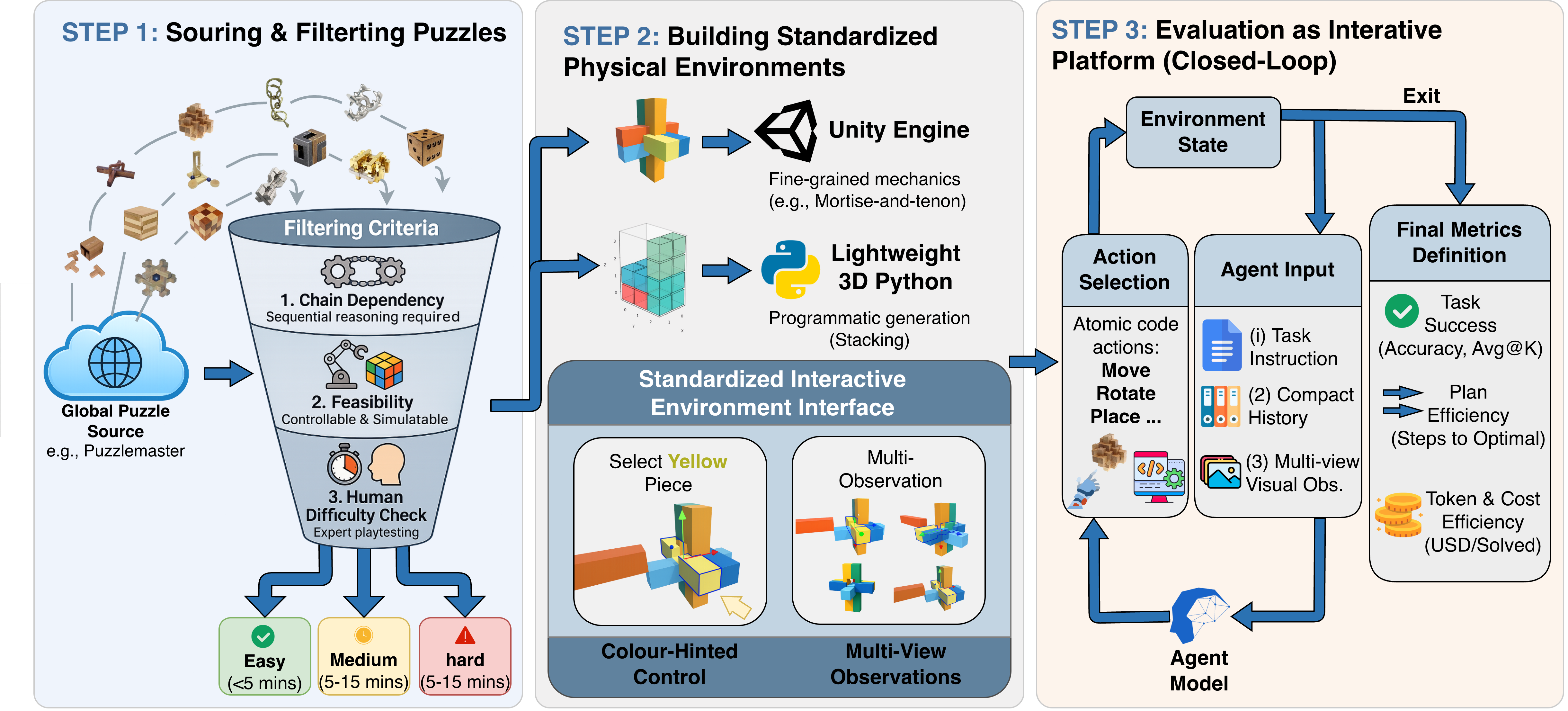}
  \caption{
  \textbf{Benchmark construction pipeline.}
  We illustrate the end-to-end process for building our benchmark, including problem sourcing,
  document collection and filtering, concept annotation, regime construction, and final evaluation setup.
  The pipeline is designed to ensure controlled difficulty, minimize parametric leakage,
  and enable fine-grained analysis of reasoning and retrieval behaviors.
  }
  \label{fig:benchmark-build}
  \vspace{-10pt}
\end{figure*}

As illustrated in \Cref{fig:benchmark-build}, we build our interactive benchmark using a three-step pipeline that ensures: (i) the puzzles are intrinsically multi-step and require solutions with sequential causal dependencies; (ii) the environments are controllable, reproducible, and of moderate difficulty; and (iii) the evaluation faithfully captures both correctness and efficiency, going beyond simple success metrics.

\paragraph{Step 1: Finding Suitable Puzzles.}
We begin by sourcing candidate puzzles from Puzzlemaster\footnote{\url{https://www.puzzlemaster.ca/}} and filtering them using three criteria.
First, we assess \textbf{chain dependency}. We retain only puzzles that require \emph{sequential} reasoning and ordered action execution, where intermediate steps must be performed in a correct order and early actions causally influence the outcomes of later ones.
Second, we evaluate \textbf{feasibility}. This includes whether a puzzle can be reliably modelled with a controllable action space and stable state transitions, and whether its difficulty lies within a regime that is both simulatable and informative for model evaluation.
Third, we conduct a \textbf{human difficulty check}. Human experts\footnote{We recruit two human experts, who are administrators of the corresponding puzzle interest groups in the forum. Each expert independently solves every candidate puzzle; we record their completion times and use the average solve time to assign the difficulty label.}
 play each candidate puzzle while we record completion time and assign difficulty labels accordingly. Based on solve time, puzzles are categorized as \textit{easy} ($<5$ minutes), \textit{medium} (5--15 minutes), or \textit{hard} ($>15$ minutes). Stacking tasks are automatically generated by code rather than curated from external sources; we nevertheless use the same human solve-time protocol to assign difficulty tiers.


\paragraph{Step 2: Building Physical Environments.}
To ensure experimental consistency, we unify the collected puzzles, spanning diverse structures and dynamics, into standardized interactive environments.
We implement these environments using two complementary toolchains: \textbf{Unity} and a lightweight \textbf{3D Python} engine.
Unity is used for puzzles with complex interlocking mechanics (e.g., Kongming and Lu Ban locks), where precise control over kinematic constraints and contact interactions is required.
In contrast, stacking-based spatial packing tasks, which involve simpler physical dynamics, are implemented in 3D Python for greater development efficiency.
To provide a uniform interaction interface across tasks, we adopt a \textbf{color-hinted control scheme}: each object is assigned a distinct color, and the color–object mapping is exposed to the agent as additional metadata. The agent specifies objects by color when selecting and manipulating pieces.
This design avoids introducing an additional action controller (as in VLA-style setups) that could confound the evaluation.
Finally, we provide \textbf{multi-view observations}, allowing agents to inspect the scene from different viewpoints and reducing failures due to occlusion.


We evaluate each instance under a closed-loop protocol that treats the benchmark as an \textbf{interactive platform}, rather than a single-turn question answering setting. For each evaluation episode, we select one task instance from either the \textbf{puzzle} suite or the \textbf{stacking} suite and \textbf{initialize the corresponding environment} to its predefined initial state (including object poses, constraints, and the fixed action set). The agent then interacts with the environment over multiple steps. At step $t$, it receives (i) a \textbf{task instruction} specifying the goal state, (ii) a compact \textbf{interaction history} summarizing prior observations and actions, and (iii) the current \textbf{multi-view visual observations} rendered by the environment. Conditioned on these inputs, the agent selects an action from the predefined action space; the simulator executes the action to \textbf{update the environment state} and returns new observations for step $t{+}1$. This perception--action loop continues until the agent solves the instance or a predefined \textbf{step budget} is reached. After the episode ends, we \textbf{compute evaluation metrics from the full trajectory}, including both the final success signal and the recorded interaction trace. Beyond task success, we also measure how efficiently the agent achieves the goal, which motivates the metric design described below. 

\subsection{Metric}
\label{subsec:metric}

We evaluate agents from three complementary perspectives: \textit{(i) task success}, \textit{(ii) plan efficiency}, and \textit{(iv) cost efficiency}. 
Together, they measure not only whether the agent solves the task, but also how efficiently it executes actions and how expensive the interaction is under realistic billing.

\paragraph{Notation.}
\providecommand{\ind}{\mathbb{1}} 
\providecommand{\Steps}{\mathrm{steps}}
\providecommand{\TokIn}{\mathrm{tokens\_in}}
\providecommand{\TokOut}{\mathrm{tokens\_out}}

We evaluate $N$ tasks (indexed by $i$). Each task yields a binary outcome $s_i\in{0,1}$ (1 if solved, 0 otherwise). During interaction, the agent executes atomic code actions, and we denote the number of executed actions by $\Steps_i\in\mathbb{N}$. For each task, we also compute a task-specific minimal plan length $o_i\in\mathbb{N}$ (i.e., the shortest valid solution plan under the environment rules). For best-of-$K$ evaluation, we sample $K$ independent runs ${a_{i,j}}_{j=1}^K$ per task.
We use $\ind(\cdot)$ to denote the indicator function.

\paragraph{(1) Task success.}
Task success measures how often an agent correctly completes a task per our evaluation protocol.

\textbf{Pass@1} (single-attempt success rate) reports the fraction of tasks that are solved in a single run:
\begin{equation}
\text{Pass@1}=\frac{1}{N}\sum_{i=1}^{N}\ind{s_i=1}
\end{equation}
where we evaluate $N$ task instances indexed by $i$, and $s_i\in{0,1}$ indicates whether the agent successfully solves task $i$ in its only attempt. Each instance therefore contributes $1$ if solved and $0$ otherwise, and $Pass@1$ is simply the mean success indicator across the benchmark.





\paragraph{(2) Plan efficiency (conditioned on success).}
Task success rate alone can mask inefficient behaviors such as redundant actions, detours, or trial-and-error. To disentangle efficiency from task difficulty, we evaluate plan efficiency \emph{only on solved tasks}. Let $\mathcal{S}=\{\,i:\ s_i=1\,\}$ denote the set of solved tasks.

\textbf{Average Steps} measures the length of successful executions (mean number of actions over solved tasks):
\begin{equation}
\text{AvgSteps}_{\text{solved}}=\frac{1}{|\mathcal{S}|}\sum_{i\in\mathcal{S}} \Steps_i
\end{equation}
For instance, if a task has an optimal length $o_i=3$ but the agent solves it using $6$ steps, it contributes $6$ to $\text{AvgSteps}_{\text{solved}}$.

\textbf{Distance-to-Optimal} quantifies avoidable overhead by counting extra steps beyond the minimal plan:
\begin{equation}
\text{Dist2Opt}=\frac{1}{|\mathcal{S}|}\sum_{i\in\mathcal{S}} \max\bigl(0,\ \Steps_i-o_i\bigr)
\end{equation}
This compares each successful trajectory against the task-specific optimum, thereby separating inefficiency from inherent task difficulty. For example, solving a $3$-step puzzle in $5$ steps adds $2$ to $\text{Dist2Opt}$.

\textbf{Normalized Distance} provides a scale-free measure by normalizing the overhead by $o_i$, making tasks with different optimal lengths comparable:
\begin{equation}
\text{NormDist}=\frac{1}{|\mathcal{S}|}\sum_{i\in\mathcal{S}} \frac{\max(0,\ \Steps_i-o_i)}{\max(1,\ o_i)}
\end{equation}
For example, a $10$-step optimal plan solved in $15$ steps yields $0.5$, whereas a $3$-step plan solved in $4$ steps yields $\tfrac{1}{3}$.

\paragraph{(3) Token and cost efficiency.}
Beyond success and action efficiency, we measure how \emph{cost-effective} an agent is to deploy in terms of both tokens and dollars.
We count all billable tokens under the target tokenizer, including prompts, tool schemas, tool inputs/outputs, and model replies, aggregated over all interaction rounds.
Let $\TokIn(i)$ and $\TokOut(i)$ denote the total input and output tokens used on task $i$, and let $s_i\in\{0,1\}$ indicate whether task $i$ is solved. To summarize token efficiency, we report Solved/1M Tokens:
{\small
\begin{equation}
\text{Solved/Tokens}
=\frac{\sum_{i=1}^{N} s_i}{\sum_{i=1}^{N}\bigl(\TokIn(i)+\TokOut(i)\bigr)}\times 10^6
\end{equation}
}

This metric rewards agents that solve more tasks while consuming fewer tokens, capturing differences in interaction length and verbosity. To translate token usage into monetary cost, we use provider prices $p_{\text{in}}$ and $p_{\text{out}}$ (USD per 1K tokens) and define the cost of task $i$ as
\begin{equation}
\text{Cost}(i)=\frac{p_{\text{in}}\cdot \TokIn(i)+p_{\text{out}}\cdot \TokOut(i)}{1000}
\end{equation}
We then report Solved/(1 USD):
\begin{equation}
\text{Solved/USD}
=\frac{\sum_{i=1}^{N} s_i}{\sum_{i=1}^{N}\text{Cost}(i)}
\end{equation}
In practice, these measures quantify the deployment trade-off between lightweight agents and larger models: similar task success can correspond to very different token footprints and therefore substantially different costs.


\section{Experiments}
    \subsection{Experimental Setup}

We evaluate a diverse set of state-of-the-art models under a unified decoding and interaction protocol to ensure fair comparison in the interactive physical reasoning setting of \bench. Concretely, our evaluated pool includes 1) closed-source models: \textit{GPT5.2}\cite{singh2025openaigpt5card}, \textit{Openai-o3}\cite{openai2024openaio1card}, \textit{Claude-Opus-4.5}\cite{anthropic2025claudeopus45systemcard}, \textit{Claude-Sonnet-4.5}\cite{anthropic2025claudesonnet45systemcard}, \textit{Gemini-3-Pro (prev.)}\cite{deepmind2025gemini3pro_modelcard}, \textit{Gemini-3-Flash (prev)}\cite{deepmind2025gemini3flash_modelcard}, \textit{Seed-1.6-Flash} and \textit{Seed-1.6}\cite{seed2025seed1}; 2) open-source models: \textit{Qwen3-VL-235B-Ins}, \textit{Qwen3-VL-30B-A3B-Ins}, \textit{Qwen3-VL-235B-Thinking}, \textit{Qwen3-VL-30B-A3B-Thinking}, \textit{Qwen3-VL-8B-Thinking}, \textit{Qwen3-VL-4B-Thinking} \cite{bai2025qwen3vltechnicalreport}, \textit{GLM-4.6V} \cite{vteam2026glm45vglm41vthinkingversatilemultimodal} and \textit{Kimi-k2.5}\cite{kimiteam2025kimik2openagentic}. (details provided in Appendix \ref{app:model_list}, spanning both general-purpose VLMs and recent strong open/closed models.

We use identical generation hyperparameters across all models: temperature is fixed to $0.6$, and nucleus sampling is set to $\text{top-}p=0.95$. For each task instance, the model interacts with the environment through the same action API, and all atomic actions are executed by the system proxy to factor out low-level control and isolate operation-level decision making. We additionally cap the interaction budget at $30-60$ steps per instance, which is substantially larger than the optimal solution length for all tasks, ensuring that failures are not driven by insufficient rollout horizon. Finally, we set the trajectory history window to $5$ (i.e., the model conditions on only the most recent five interaction turns when making each decision).

\subsection{Main Experiment}

\begin{table*}[t]
\centering
\footnotesize
\renewcommand{\arraystretch}{1.15}
\setlength{\tabcolsep}{4.6pt}
\resizebox{1.0\textwidth}{!}{
\begin{tabular}{l|rrrr|rrr|rr}
\toprule
\multirow{2.5}{*}{\textbf{Models}}
& \multicolumn{4}{c|}{\textbf{Task Success}}
& \multicolumn{3}{c|}{\textbf{Plan Efficiency (Solved)}}
& \multicolumn{2}{c}{\textbf{Token \& Cost}} \\
\cmidrule(lr){2-5}\cmidrule(lr){6-8}\cmidrule(lr){9-10}
& \textbf{Pass@1} $\uparrow$
& \textbf{Succ.Task} $\uparrow$
& \textbf{Puzzle} $\uparrow$
& \textbf{Stacking} $\uparrow$
& \textbf{AvgSteps} $\downarrow$
& \textbf{Dist2Opt} $\downarrow$
& \textbf{N.Dist} $\downarrow$
& \textbf{Solved/Tokens} $\uparrow$
& \textbf{Solved/USD} $\uparrow$ \\
\midrule

\multicolumn{10}{l}{\cellcolor{lightblue}\textit{Closed-source Models}} \\
\midrule
GPT-5.2 & \textbf{22.9} & \textbf{25} & 3.1 & \textbf{31.2} & 9.6 & 1.43 & 0.50 & 10.1 & 0.10 \\
GPT-5-mini & 11.0 & 12 & 3.1 & 14.3 & 6.8 & 1.08 & 0.57 & 19.3 & 0.19 \\
OpenAI-o3 & 10.1 & 11 & 3.1 & 13.0 & 6.1 & 0.81 & 0.37 & 26.1 & 0.26 \\
Claude-Opus-4.5 & 15.6 & 17 & 3.1 & 20.8 & 8.0 & 1.42 & 0.63 & 17.8 & 0.17 \\
Claude-Sonnet-4.5 & 13.8 & 15 & 3.1 & 18.2 & 7.1 & 1.02 & 0.51 & 29.0 & 0.29 \\
Gemini-3-Pro & 19.3 & 21 & 3.1 & 26.0 & 10.4 & 2.38 & 1.29 & 4.63 & 0.04 \\
Gemini-3-Flash & 11.9 & 13 & 3.1 & 15.6 & 6.1 & 0.84 & 0.46 & 36.9 & 0.36 \\
Seed-1.6 & 10.1 & 11 & 3.1 & 13.0 & 5.5 & 0.73 & 0.39 & 23.9 & 0.23 \\
Seed-1.6-Flash & 7.3 & 8 & 0.0 & 10.4 & 4.5 & 0.23 & 0.07 & 32.4 & 0.32 \\

\midrule
\multicolumn{10}{l}{\cellcolor{mediumblue}\textit{Open-source Models}} \\
\midrule
Kimi-k2.5 & 13.8 & 15 & 0.0 & 19.5 & 7.0 & 0.71 & 0.19 & 6.2 & 0.06 \\
GLM-4.6V & 7.3 & 8 & 0.0 & 10.4 & 6.5 & 0.92 & 0.39 & 22.6 & 0.22 \\
Qwen3-VL-235B-A22B-Thk. & 9.2 & 10 & 3.1 & 11.7 & 3.8 & 0.33 & 0.31 & 40.7 & 0.40 \\
Qwen3-VL-30B-A3B-Thk. & 10.1 & 11 & 0.0 & 14.3 & 5.5 & 0.40 & 0.11 & 18.8 & 0.18 \\
Qwen3-VL-8B-Thinking & 8.3 & 9 & 0.0 & 11.7 & 3.8 & \textbf{0.03} & \textbf{0.01} & 19.8 & 0.19 \\
Qwen3-VL-235B-A22B-Inst & 8.3 & 9 & 3.1 & 10.4 & \textbf{3.7} & 0.13 & 0.11 & \textbf{66.8} & \textbf{0.66} \\
Qwen3-VL-30B-A3B-Inst & 3.7 & 4 & 3.1 & 3.9 & 4.8 & 1.30 & 1.14 & 51.4 & 0.51 \\

\bottomrule
\end{tabular}}
\caption{
Main model evaluation results. We report overall task success on the full benchmark (Pass@1 (All)) and subset success on Puzzle \& Stacking. Plan-efficiency metrics (AvgSteps, Dist2Opt, NormDist; lower = better) are computed only for solved tasks. Token and monetary costs (Solved/Tokens, Solved/USD) are normalized by successful tasks.
}
\label{tab:main-results}
\vspace{-8pt}
\end{table*}

\label{main-result}

Table~\ref{tab:main-results} reports the performance of a broad set of closed- and open-source models on \bench~across all task families and difficulty levels. Overall, closed-source models dominat the leaderboard. GPT-5.2 achieves the best aggregate result (Pass@1=22.9, Succ~Task=25) and also attains the highest \textit{Stacking} score (31.2). Among open-source models, Kimi-k2.5 is the strongest overall (Pass@1=13.8, Succ~Task=15). Despite steady gains, \bench~remains challenging: the \textbf{3D puzzle is a major bottleneck} for all evaluated VLMs. In particular, \textit{Puzzle} success is consistently far below \textit{Stacking} (most models reach only \(\approx 0.0\)–\(3.1\) on \textit{Puzzle}, versus \(10.4\)–\(31.2\) on \textit{Stacking}). This gap suggests that even state-of-the-art models struggle with the structural inference and constrained, sequential manipulation required by the hardest human-designed 3D puzzles. Qualitatively, many models fail early and \emph{stall around the second level}, indicating that failures are driven less by minor execution noise and more by difficulty in identifying valid intermediate constraints.

\begin{figure}
    \centering
    \includegraphics[width=\linewidth]{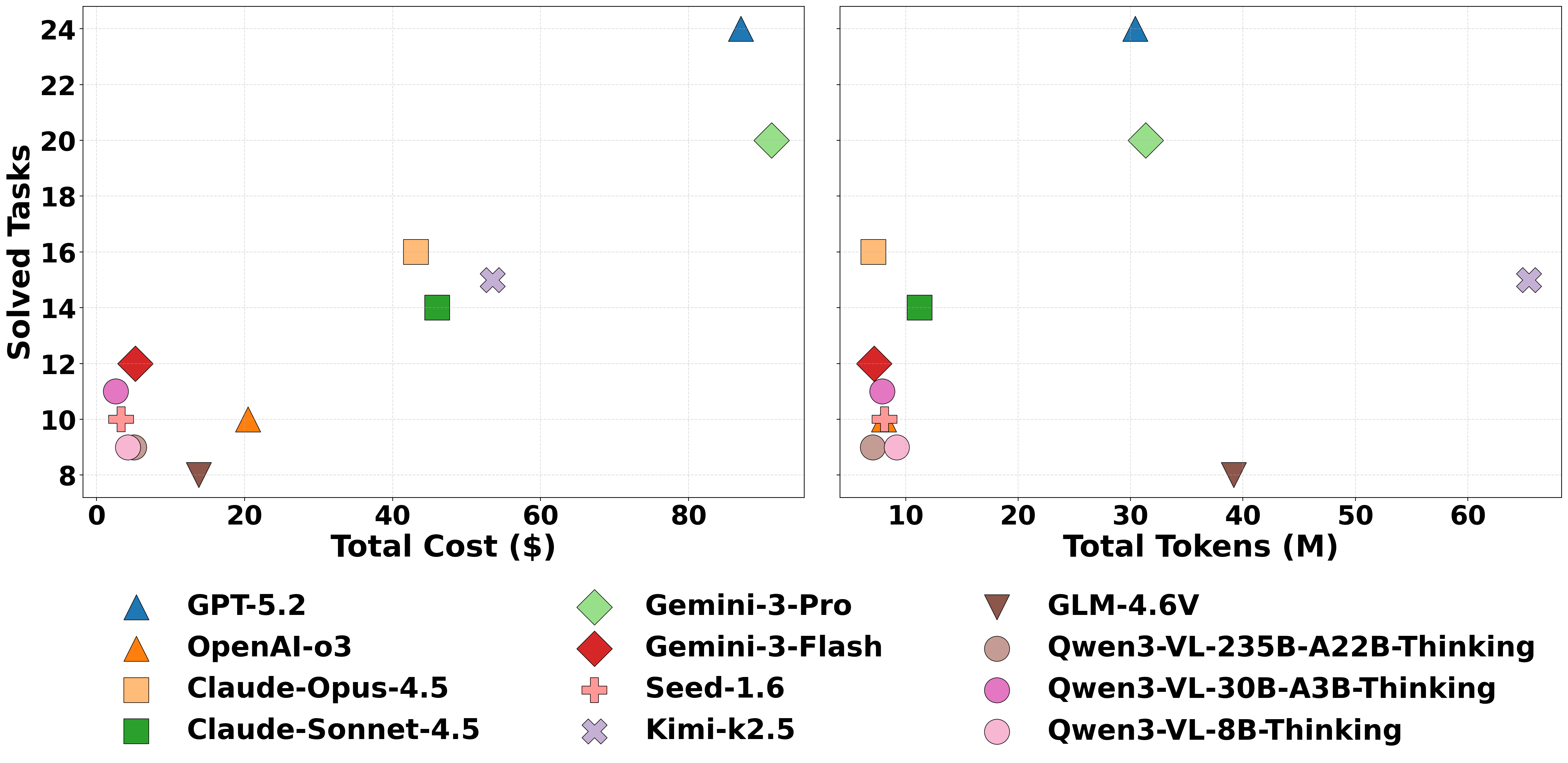}
    \caption{Cost and token efficiency with solved tasks comparison between models}
    \label{fig:cost_plot}
    \vspace{-20pt}
\end{figure}

Higher success does not always mean better efficiency or lower cost. Stronger models often \emph{backtrack and revise} their plans, which increases AvgSteps/Dist2Opt/NormDist and raises Solved/Tokens and Solved/USD. For instance, GPT-5.2 is relatively expensive per solved task (\$1.3/\text{level},)). In contrast, lightweight “flash” models can be extremely cheap  but achieve much lower success. To visualize the cost--quality trade-off, Fig.~\ref{fig:cost_plot} plots per-solved cost and token usage on the $x$-axis against overall success on the $y$-axis; points closer to the \emph{upper-left} represent better value. Notably, in \emph{easy} regimes, extremely small models are not always the most cost-effective choice: low success can inflate cost-per-success, whereas a moderately stronger model may solve more reliably with fewer failed attempts, yielding a better effective return despite higher per-call expense.

Finally, we analyze representative failure cases to pinpoint what most limits progress on each task family. For \textit{Puzzle}, the primary obstacle is weak perception and utilization of \emph{physical structure}: even when partial internal structure is revealed as a reference, many models still cannot reliably identify the \emph{first critical move} that unlocks the puzzle. As a result, they resort to extensive trial-and-error over candidate blocks, leading to largely random exploration with little constraint-guided progress. For \textit{Stacking}, two factors dominate failures. First is \emph{object-set complexity and coupling}: as illustrated in Fig.~\ref{fig:stacking_level_examples}, while easy instances can be solved with locally reasonable placements, mid and hard levels require tightly coupled packing decisions to maximize space utilization. Second is \emph{global spatial planning}: most models greedily place “easy” items early, only to later discover that the remaining free volume is fragmented, forcing costly removals and replanning. Stronger models such as GPT-5.2 and Gemini-Pro exhibit more deliberate lookahead, which helps avoid these dead-ends and explains a substantial portion of their advantage on difficult stacking regimes. For a specific discussion of the factors contributing to success and failure, please refer to Appendix~\ref{app:case_analysis}.

\begin{figure*}
    \centering
    \includegraphics[width=0.75\linewidth]{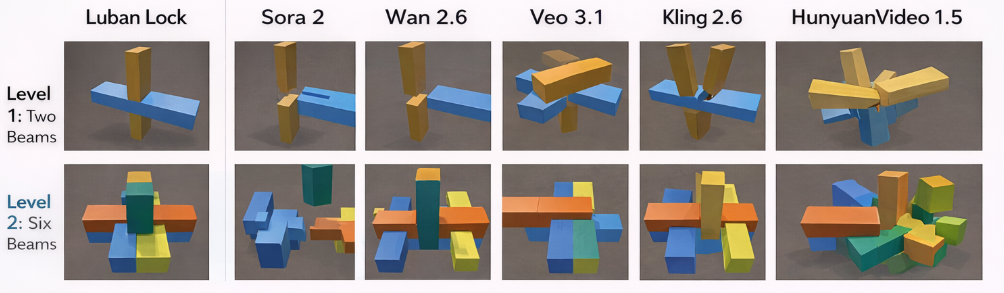}
    \caption{
Qualitative results on the Luban puzzle subtask across world models.
Top: \textbf{Level 1} (two beams). Bottom: \textbf{Level 2} (six beams).
All models fail to produce a physically valid disassembly, either violating interlocking constraints or hallucinating (e.g., structural corruption and object insertion/removal), with failures worsening at higher complexity.
}
    \label{fig:case_world_model}
    \vspace{-10pt}
\end{figure*}

\subsection{Catastrophic Failure of World Models}
\label{sec:catastrophic_world_model}

To further extend physical interaction into the setting of world models, we use our \textbf{puzzle subtask} to assess world models for producing \textbf{step-by-step disassembly process} under explicit physical constraints. Given a reference image of a Luban lock, models are instructed to disassemble all pieces while respecting contact, interlocking, and collision rules (see the detailed prompt in Table~\ref{app:world_model_prompt}). We evaluate state-of-the-art video generation models (\textsc{Sora 2}\cite{Li2025Sora2}, \textsc{Wan 2.6}\cite{wan2025wanopenadvancedlargescale}, \textsc{Veo 3.1}\cite{wiedemer2025videomodelszeroshotlearners}, \textsc{Kling 2.6}\cite{Kuaishou2025Kling26}, and \textsc{HunyuanVideo 1.5}\cite{hunyuanvideo2025}) on two configurations: \textbf{Level 1} (two beams) and \textbf{Level 2} (six beams). Despite the highly specified prompts, none of the models successfully completes the disassembly; instead, they exhibit systematic, catastrophic failures with severe hallucinations shown in Figure~\ref{fig:case_world_model}, which become more pronounced as structural complexity increases.

First, some models exhibit \textbf{superficial instruction-following with physics violations}. For example, \textsc{Sora 2} and \textsc{Wan 2.6} often extract a target beam via direct translation, even when such motion is physically infeasible due to interlocking constraints. It further degrades in Level 2, where these models increasingly fail to follow the prescribed step-by-step procedure and instead perform random or under-specified actions that remain physically invalid.

Second, other models undergo a more severe \textbf{representational collapse}, losing consistency of both object structure and object identity. In Level 1, \textsc{Veo 3.1}, \textsc{Kling 2.6}, and \textsc{HunyuanVideo 1.5} frequently generate corrupted configurations, including distorted geometry and spurious components. In Level 2, this behaviour escalates into full hallucination: models may add, remove, or merge beams, or transform the puzzle into an unrecognisable structure. Such failures indicate a breakdown of object permanence and constraint consistency under multi-part coupling.

Overall, these results demonstrate that although modern world models can generate visually plausible motion or simple physical event transitions, they remain \textbf{fundamentally unreliable for structured, constraint-driven interaction}. In particular, tasks requiring multi-step manipulation grounded in object-centric reasoning and physical feasibility remain beyond the capabilities of current models, highlighting an important direction for future world model research.

\subsection{Impact of Difficulty Stratification}

\begin{table}[t]
\centering
\footnotesize
\renewcommand{\arraystretch}{1.15}
\setlength{\tabcolsep}{4.2pt}
\resizebox{0.95\linewidth}{!}{
\begin{tabular}{l|ccc|ccc}
\toprule
\multirow{2.2}{*}{\textbf{Models}}
& \multicolumn{3}{c|}{\textbf{Puzzle Acc} $\uparrow$}
& \multicolumn{3}{c}{\textbf{Stacking Acc} $\uparrow$} \\
\cmidrule(lr){2-4}\cmidrule(lr){5-7}
& \textbf{Easy} & \textbf{Mid} & \textbf{Hard}
& \textbf{Easy} & \textbf{Mid} & \textbf{Hard} \\
\midrule
GPT5.2               & 10.0 & 0.0 & 0.0 & 100.0 & 55.0 & 6.3 \\
Gemini-3-Pro         & 10.0 & 0.0 & 0.0 & 90.0 & 40.0 & 6.3 \\
Sonnet-4.5    & 10.0 & 0.0 & 0.0 & 100.0 & 20.0 & 0.0 \\

\bottomrule
\end{tabular}}
\caption{
Difficulty-stratified accuracy (Acc) on \bench{}.
We report Easy/Mid/Hard tiers for both Puzzle and Stacking.
}
\vspace{-20pt}
\label{tab:difficulty-stratified-acc}
\end{table}

Difficulty in Stacking increases with both the number of blocks and the complexity of support relations, making success increasingly dependent on long-horizon planning. This is reflected in Table~\ref{tab:difficulty-stratified-acc}: Stacking-Easy is essentially solved by top models (e.g., GPT5.2 and Claude-Sonnet-4.5 at 100.0\%), but the gap opens on Stacking-Mid (e.g., \textit{GPT5.2} at 55.0\% vs.\ \textit{Claude-Sonnet-4.5} at 20.0\%) and performance collapses on Stacking-Hard (best results only 6.3\%).

In contrast, Puzzle is consistently harder across tiers: even on Puzzle-Easy models peak at only 10.0\%, and Puzzle-Mid/Hard remains at 0.0\%. This suggests the dominant bottleneck is not a smooth easy-to-hard degradation, but the fundamentally higher difficulty of 3D interlocking/structure-centric reasoning itself---requiring inference of hidden blocking constraints and feasible multi-step disentanglement trajectories from partial observations\footnote{A more detailed description of the difficulty-level definitions can be found in Appendix~\ref{app:task_case}}.

\subsection{One-shot Solving without Interaction}

\begin{table}[t]
\centering
\footnotesize
\renewcommand{\arraystretch}{1.15}
\setlength{\tabcolsep}{4.6pt}
\resizebox{0.98\linewidth}{!}{
\begin{tabular}{l|ccc|ccc|c}
\toprule
\multirow{2.2}{*}{\textbf{Models}}
& \multicolumn{3}{c|}{\textbf{Interactive (\%) $\uparrow$}}
& \multicolumn{3}{c|}{\textbf{One-shot (\%) $\uparrow$}}
& \multirow{2.2}{*}{\textbf{$\Delta$}} \\
\cmidrule(lr){2-4}\cmidrule(lr){5-7}
& \textbf{Puzzle} & \textbf{Stacking} & \textbf{All}
& \textbf{Puzzle} & \textbf{Stacking} & \textbf{All}
& \\
\midrule
GPT5.2               & 3.1 & 31.2 & 22.9 & 0.0 & 9.1 & 7.1 & -15.8 \\
Sonnet-4.5    & 3.1 & 18.2 & 13.8 & 0.0 & 10.3 & 8.1 & -5.7 \\
Gemini-3-Pro         & 3.1 & 26.0 & 19.3 & 0.0 & 9.1 & 7.1 & -12.2 \\
\bottomrule
\end{tabular}}
\caption{
Interactive vs.\ one-shot solving on \bench{}.
In one-shot, the model receives only a single fixed-view image and must output a complete (or near-complete) solution with minimal or no intermediate feedback.
$\Delta$ All Avg denotes the performance gap between interactive and one-shot settings (Interactive $-$ One-shot).
}
\label{tab:oneshot-vs-interactive}
\vspace{-10pt}
\end{table}

A key motivation for interactive evaluation is to distinguish \emph{offline planning} from \emph{closed-loop adaptation}.
In physical problem solving, interaction can compensate for imperfect priors: a model may probe the environment, observe feasibility constraints, and revise its plan accordingly.
To isolate how much success on \bench{} depends on such feedback, we construct a one-shot setting where the model receives only a single fixed-view image and must output a complete solution with minimal or no intermediate feedback.
We then compare one-shot accuracy with the interactive setting; the gap $\Delta$ (Interactive $-$ One-shot) quantifies the benefit brought by interaction.

Table~\ref{tab:oneshot-vs-interactive} shows that one-shot performance is uniformly lower, which further highlights gains of interactions.
This indicates that \bench{} cannot be reliably solved by pre-computed reasoning from a single view.
For \textbf{Puzzle}, one-shot accuracy collapses to $0.0\%$ for all evaluated models, while interactive accuracy reaches $3.1\%$, suggesting that even modest success requires iterative constraint discovery rather than a fully inferred disassembly plan from the initial observation.
For \textbf{Stacking}, interaction is even more critical: GPT5.2 drops from $31.2\%$ (interactive) to $9.1\%$ (one-shot), Gemini-3-Pro shows the same drop ($26.0\%\rightarrow 9.1\%$), and Claude-Sonnet-4.5 decreases from $18.2\%$ to $10.3\%$.
Aggregated over all tasks, the All Avg accuracy decreases sharply in one-shot.
Overall, these consistent gaps support that \bench{} genuinely evaluates closed-loop physical-structure reasoning under evolving feasibility constraints, rather than one-shot recognition or static plan synthesis.

\subsection{Reward Models vs. Verifier-based Checking}

\begin{table}[t]
\centering
\footnotesize
\renewcommand{\arraystretch}{1.15}
\setlength{\tabcolsep}{4.8pt}
\resizebox{1.0\linewidth}{!}{
\begin{tabular}{l|ccc|c}
\toprule
\textbf{Strategy}
& \textbf{Puzzle(\%) $\uparrow$}
& \textbf{Stacking(\%) $\uparrow$}
& \textbf{All(\%) $\uparrow$}
& \textbf{$\Delta$ vs.\ Avg@4 $\uparrow$} \\
\midrule
Avg@4           & 3.1  & 15.5   & 9.3  & -- \\
\midrule
Pass@1              & 3.1 & 15.6 & 9.4 & +0.1 \\
Pass@2              & 3.1  & 19.4   & 11.2  & +1.9 \\
Pass@4              & 3.1  & 19.4   & 11.2  & +1.9 \\
\midrule
VLM judge      & 3.1  & 18.1   & 10.3  & +1.3 \\
Reward model          & 3.1  & 16.8   & 9.9  & +0.6 \\
\bottomrule
\end{tabular}}
\caption{
Comparison of multi-sample selection and reward-model (RM) reranking on \bench{} by kimi-k2.5.
All strategies use the same base generator and differ only in candidate selection.
$\Delta$ is computed w.r.t.\ Avg@4 on All Avg.
}
\label{tab:rm-vs-verifier}

\vspace{-15pt}
\end{table}

RLVR indicates verification greatly aids domains with deterministic checking (e.g., math/code). We question if this applies to interactive physical reasoning—where success relies on long action sequences and errors may emerge late. Supervision is tougher here: reward models often favor locally plausible moves, whereas verifier-style checks are more conservative but environment-based.

We therefore compare RM-based selection\cite{ong2025trainingvisionlanguageprocessreward} with verifier-based checking for choosing candidate actions/plans (e.g., execution-grounded validity/consistency), keeping the generator fixed, and testing lightweight strategies like reranking and pairwise judging and using Kimi-k2.5 as action model (Table~\ref{tab:rm-vs-verifier}).

Results show Pass@1 and Avg@4 are nearly identical overall, implying limited sampling variance. Multi-sample pass selection helps but saturates, while Pass@4 adds no further gain. RM reranking yields smaller improvements (+0.6), whereas a stronger VLM pairwise judge does better (+1.3) but still trails Pass@2.  Beam search (batch size=2) with Reward also cannot match Pass@2, suggesting the bottleneck is selection-signal quality rather than decoding compute. Overall, current vision RMs offer limited leverage for long-horizon interactive planning, and more reliable verifier-style signals appear necessary for consistent gains.




\section{Related Work}

\paragraph{Reasoning of Vision-Language Models.}

While VLMs effectively align visual perception with language reasoning, they remain largely confined to static scenes and single-step inference with limited temporal dynamics~\cite{sarch2025grounded}. 
Embodied extensions of these agents~\cite{liu2024moka} still primarily rely on instantaneous observations, frequently failing in multi-round physical interactions~\cite{Guo2024PhyGrasp}. Despite recent advances in multi-stage supervision and reinforcement learning~\cite{xu2411llava, guo2025deepseek, liu2025visual, kang2025viki}, these strategies are mostly validated on static benchmarks. \bench~addresses this by targeting underexamined causal and interactive reasoning over multi-step physical processes.

\paragraph{Physical Benchmarks.}

Physical reasoning evaluation has evolved from visual plausibility in simplified settings~\cite{riochet2018intphys, rajani2020esprit} to synthetic primitives in controlled environments~\cite{yi2019clevrer, wang2024compositional, zheng2024contphy, physion++}. While many benchmarks emphasize commonsense QA~\cite{lu2022learn, he2024olympiadbench, physbench2025} or broader perception-centered tasks~\cite{chow2025physbenchbenchmarkingenhancingvisionlanguage}, they remain largely static. \bench~shifts this paradigm toward dynamic, interaction-driven evaluation of multi-step causal processes.

\paragraph{3D Structure Perception.}

Foundational work in 3D structure perception~\cite{Chen_2024_CVPR, lyu2024mmscan, wang2023embodiedscan} has enabled models to reason about depth and relative positioning via multi-view or point-cloud representations~\cite{Zhao2025CoTVLAVC, lu2025scalingagenticreinforcementlearning, ma2026causalspatialbenchmarkobjectcentriccausal, wang2026visgym}. However, these efforts primarily prioritize static scene reconstruction or snapshot reasoning. Consequently, existing benchmarks rarely assess how spatial configurations evolve through interaction or how actions induce causal changes over time.

\section{Conclusion}
    

We introduced \bench~, an interactive 3D benchmark that shifts the evaluation of vision-language models from passive perception to closed-loop, multi-step physical reasoning. Across interlocking puzzles and 3D stacking/packing, models must repeatedly observe outcomes, choose feasible interactions, and revise plans as geometry, contacts, and support relations restrict what can be done next. Our evaluation of state-of-the-art VLMs and diffusion-based image-to-video models shows a clear pattern: performance degrades sharply as structural constraints tighten, and many models fail to maintain coherent multi-step strategies even when they can perceive the scene correctly. The most striking failure occurs on interlocking mechanical puzzles, where hidden geometric constraints cause near-total collapse; stacking tasks further expose brittle stability and long-horizon feasibility reasoning. These results highlight a persistent gap between “seeing” and “acting”: current models rarely internalize how early actions reshape the future feasible action space.
\newpage
\section*{Impact Statements}
This paper presents work whose goal is to advance machine learning by enabling more rigorous evaluation of interactive, physics-grounded vision reasoning through an open benchmark and reproducible baselines. There are many potential societal consequences of improved interactive perception-and-action systems, but we do not anticipate any direct negative impacts from releasing this evaluation benchmark beyond standard concerns for open research artifacts.
\newpage

\nocite{langley00}

\bibliography{example_paper}
\bibliographystyle{icml2026}


\appendix
\onecolumn
\section{Appendix}

\subsection{Limitations}
We acknowledge two main limitations of the current \textsc{Chain} benchmark. 
\textbf{(1) Limited scale of interactive environments, especially for interlocking puzzles.}
Our benchmark currently contains a finite set of \emph{Puzzle} and \emph{Stacking} instances (32 and 77, respectively). 
While stacking instances are \emph{programmatically generated} and can be scaled to near-unlimited complexity, 
high-fidelity interlocking puzzles require \emph{careful environment engineering} to faithfully capture kinematic feasibility and contact-rich constraints, and we implement them in Unity for precise control of interlocking mechanics.
As a result, each new puzzle environment incurs substantial manual modeling and debugging effort, leading to significant time and cost overhead, which currently limits benchmark scale.

\textbf{(2) Current evaluation mainly reports Pass@1 due to high multi-step interaction cost.}
We primarily report \textbf{Pass@1} (single-attempt success) as our main success metric. 
This choice is partly driven by the evaluation cost of closed-loop interaction: each episode involves a non-trivial number of rounds, and we cap the interaction budget at \textbf{30--60 steps per instance}.
We recognize that interactive tasks can exhibit run-to-run variability; however, our preliminary analysis in Section~3.6 suggests that multi-sample evaluation (e.g., Avg@4) and Pass@1 show consistent trends, indicating that the variance is not dominant in practice. 
Looking forward, once sufficient API budget is available, we will report more robust best-of-$K$ results (e.g., \textbf{Pass@4}) following the standard multi-run protocol. 

\subsection{Model list}
\label{app:model_list}
\paragraph{Evaluated Models.}
We evaluate a broad pool of state-of-the-art multimodal LLMs, covering both closed-source APIs and open-source checkpoints.
Specifically, our \textit{closed-source} set includes GPT5.2, o3, Opus4.5, Sonnet4.5, Gemini3-Pro (preview), Gemini3-Flash (preview), and Seed1.6 (with Seed1.6-Flash as an optional lightweight variant).
Our \textit{open-source} set consists of Qwen3VL-235B (A22B, think), Qwen3VL-30B (A3B, think), Qwen3VL-8B (think), Qwen3VL-2B (think), GLM4.6V, MiMo-v2-Flash, and Kimi-VL-A3B (think).
Table~\ref{tab:model-specs} summarizes each model’s release date, maximum context window, and parameter scale (when disclosed).
For MoE models, we additionally report the active parameter count per forward pass in the form of \texttt{AxxB}.
Models whose parameter sizes are not publicly released are marked as \texttt{--}.

\begin{table}[t]
\centering
\footnotesize
\renewcommand{\arraystretch}{1.15}
\setlength{\tabcolsep}{5.2pt}
\resizebox{0.7\textwidth}{!}{
\begin{tabular}{l|ccc}
\toprule
\multirow{2.2}{*}{\textbf{Models (abbr.)}}
& \multicolumn{3}{c}{\textbf{Model Specs}} \\
\cmidrule(lr){2-4}
& \textbf{Release Date}
& \textbf{Context (tokens)}
& \textbf{Model Size} \\
\midrule

\multicolumn{4}{l}{\cellcolor{lightblue}\textit{Closed-source Models}} \\
\midrule
GPT5.2                    & 2025-12-11 & 400,000   & -- \\
o3 (optional)             & 2025-04-16 & 200,000   & -- \\
Opus4.5 (optional)        & 2025-11-24 & 200,000   & -- \\
Sonnet4.5                 & 2025-09-29 & 1,000,000 & -- \\
Gemini3-Pro (preview)     & 2025-11-18 & 1,048,576 & -- \\
Gemini3-Flash (preview)   & 2025-12-17 & 1,000,000 & -- \\
Seed1.6-Flash (optional)  & 2025-12-23 & 262,144   & -- \\
Seed1.6                   & 2025-12-23 & 262,144   & -- \\

\midrule
\multicolumn{4}{l}{\cellcolor{lightblue}\textit{Open-source Models}} \\
\midrule
Qwen3VL-235B (A22B, think) & 2025-09-23 & 262,144 & 235B (A22B) \\
Qwen3VL-30B (A3B, think)   & 2025-10-06 & 131,072 & 30B (A3B) \\
Qwen3VL-8B (think)         & 2025-10-14 & 256,000 & 8B \\
Qwen3VL-2B (think)         & 2025-10-21 & 256,000 & 2B \\
GLM4.6V (optional)         & 2025-12-08 & 131,072 & 106B (A12B) \\
MiMo-v2-Flash (optional)   & 2025-12-14 & 262,144 & 309B (A15B) \\
Kimi-VL-A3B (think, opt.)  & 2025-04-10 & 131,072 & -- (A2.8B) \\
\bottomrule
\end{tabular}}
\caption{
Model release time, context window, and parameter scale for the evaluated model pool.
For MoE models, \texttt{AxxB} denotes active parameters per forward pass when available.
Unknown values are marked as \texttt{--}.
}
\label{tab:model-specs}
\end{table}

\subsection{Task Examples}
\label{app:task_case}
\begin{figure}[htbp]
  \centering

  \begin{subfigure}[b]{0.32\textwidth}
    \centering
    \includegraphics[width=\linewidth]{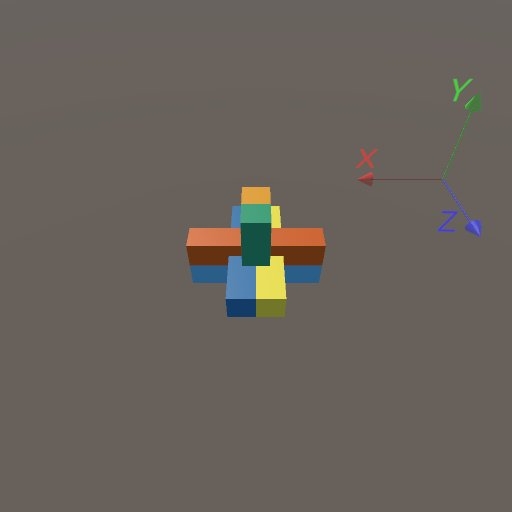}
    \caption{Easy Level}
    \label{fig:puzzle_easy}
  \end{subfigure}
  \hfill
  \begin{subfigure}[b]{0.32\textwidth}
    \centering
    \includegraphics[width=\linewidth]{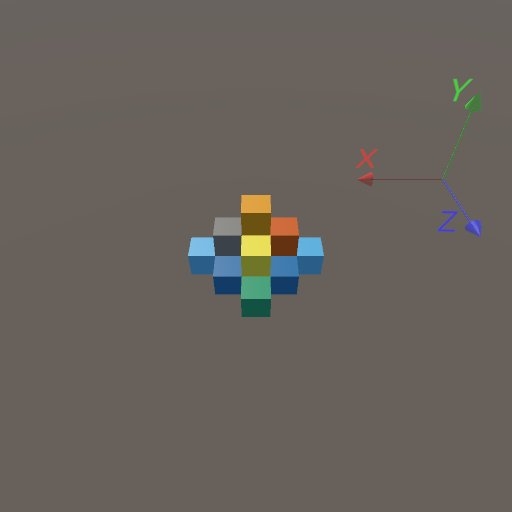}
    \caption{Middle Level}
    \label{fig:puzzle_mid}
  \end{subfigure}
  \hfill
  \begin{subfigure}[b]{0.32\textwidth}
    \centering
    \includegraphics[width=\linewidth]{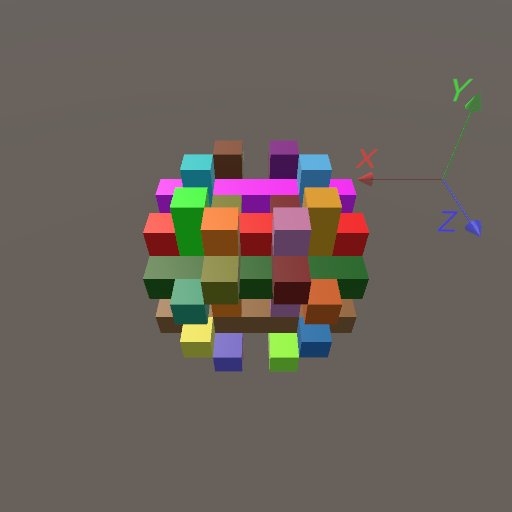}
    \caption{Hard Level}
    \label{fig:puzzle_hard}
  \end{subfigure}

  \caption{Examples of different levels of puzzle task.}
  \label{fig:puzzle_level_examples}
\end{figure}

This section showcases some cases of puzzles and stacking. As shown in Fig.~\ref{fig:puzzle_level_examples}, we divide the puzzle task into three difficulty levels according to (i) the number of pieces and (ii) the complexity of interlocks:
\begin{itemize}
    \item \textbf{Easy Level:} the object consists of a small number of pieces with simple interfaces and weak interlocking. Each piece has a distinctive placement, and the disassembly can be completed with few steps and low risk of dead-ends.
    \item \textbf{Middle Level:} the object contains more pieces and more complex contact surfaces. Interlocking features become more frequent, requiring more careful selection of the moving direction and intermediate ordering.
    \item \textbf{Hard Level:} the object consists of many pieces with strong mutual interlocks (e.g., multiple pieces jointly constrain removal), often necessitating long-horizon planning and reasoning over constraint propagation across the entire disassembly.
\end{itemize}

\begin{figure}[htbp]
  \centering

  \begin{subfigure}[b]{0.32\textwidth}
    \centering
    \includegraphics[width=\linewidth]{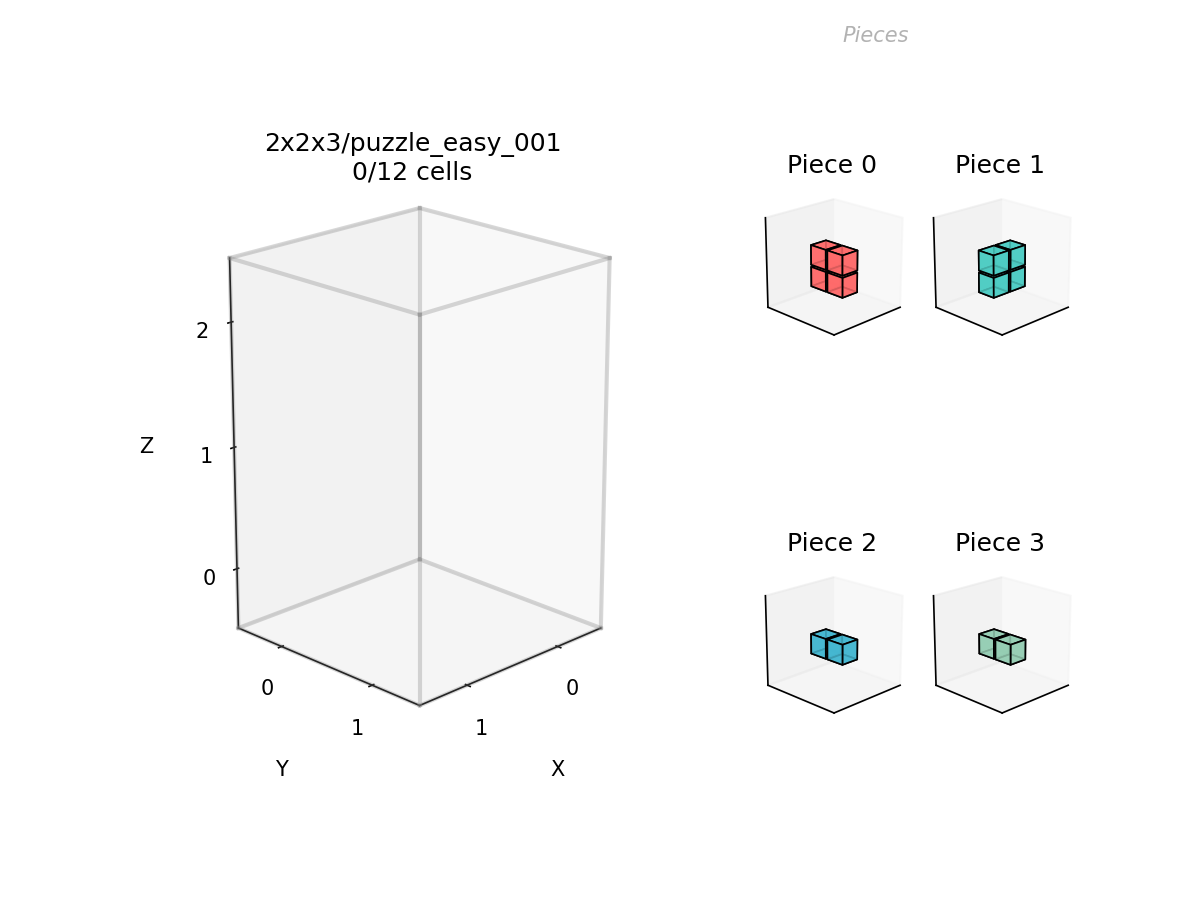}
    \caption{Easy Level}
    \label{fig:stacking_easy}
  \end{subfigure}
  \hfill
  \begin{subfigure}[b]{0.32\textwidth}
    \centering
    \includegraphics[width=\linewidth]{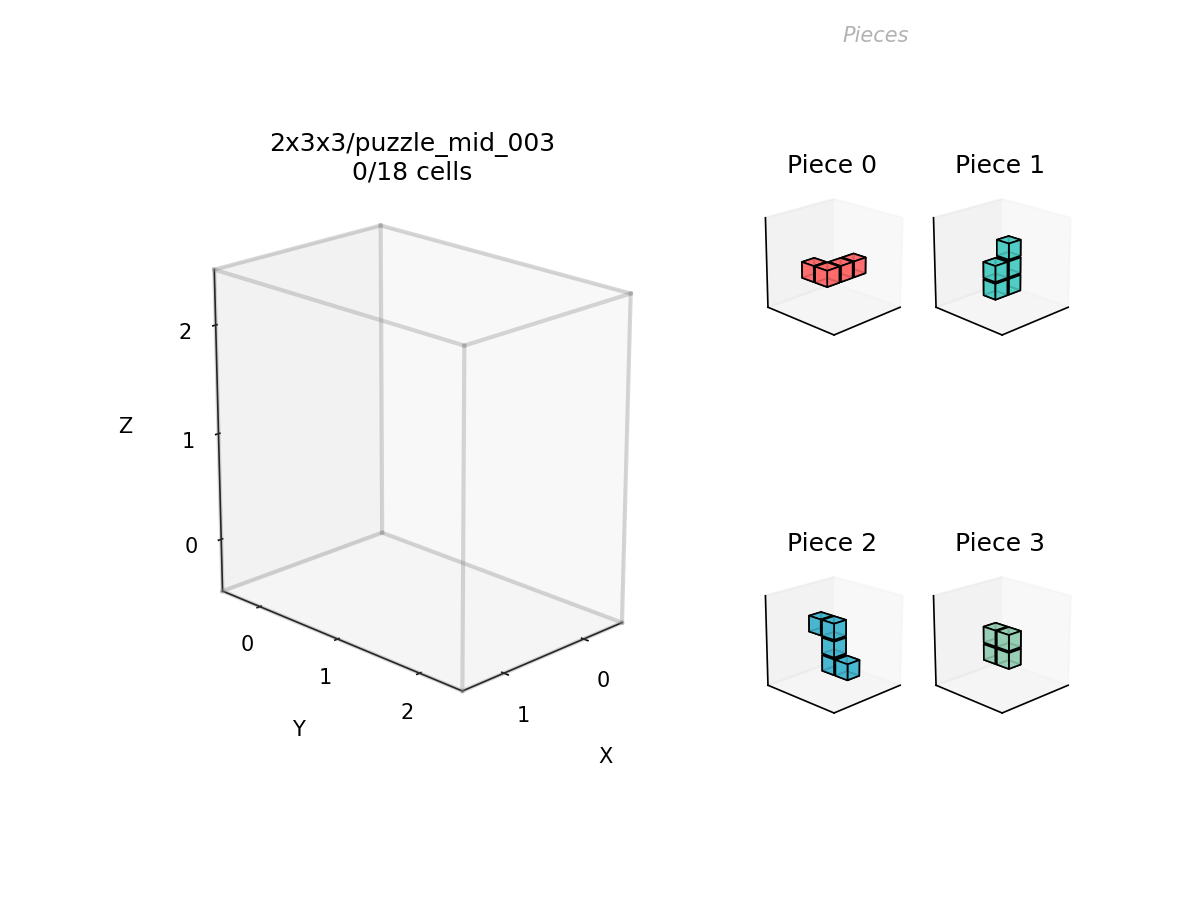}
    \caption{Middle Level}
    \label{fig:stacking_mid}
  \end{subfigure}
  \hfill
  \begin{subfigure}[b]{0.32\textwidth}
    \centering
    \includegraphics[width=\linewidth]{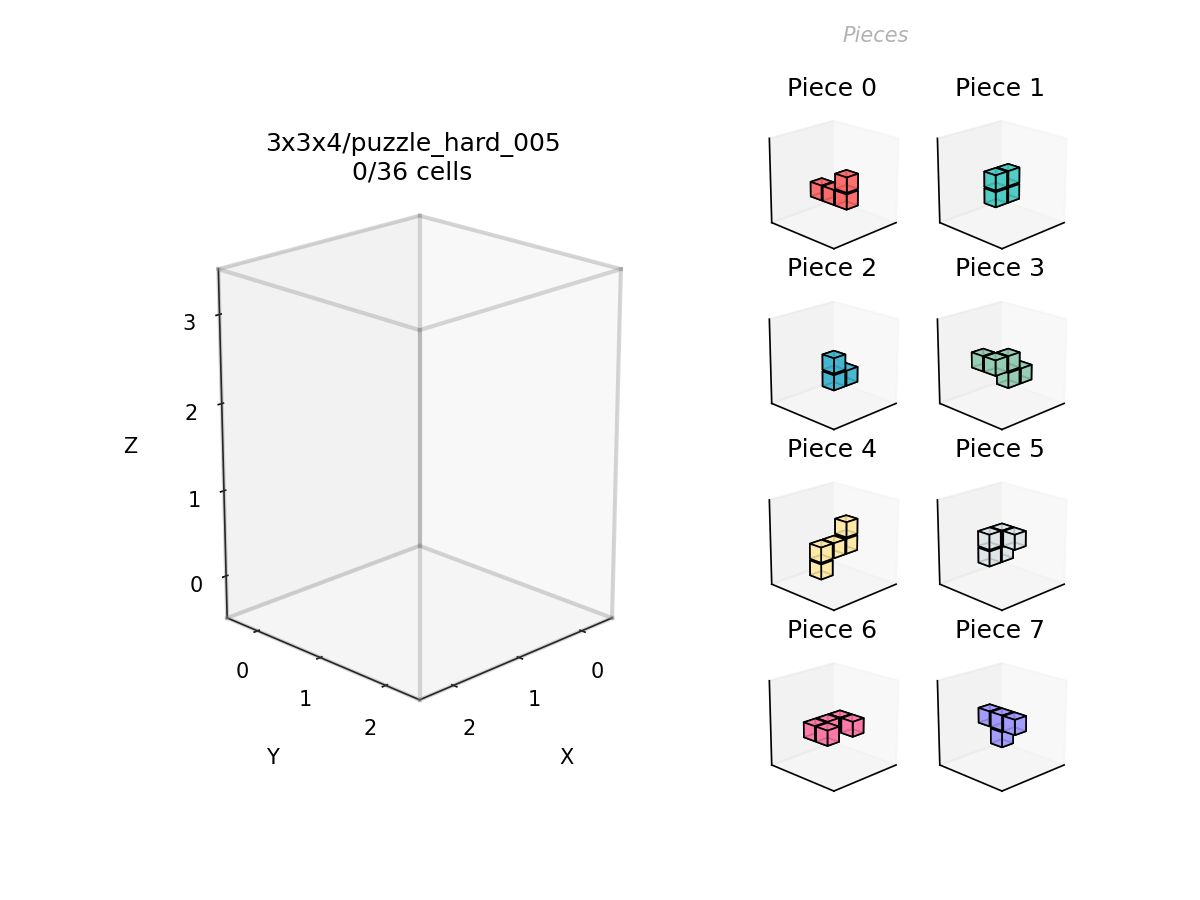}
    \caption{Hard Level}
    \label{fig:stackng_hard}
  \end{subfigure}

  \caption{Examples of different levels of stacking task.}
  \label{fig:stacking_level_examples}
\end{figure}

As shown in Fig.~\ref{fig:stacking_level_examples}, we categorize stacking difficulty primarily by the container size and the number and shape diversity of pieces, which jointly determine the combinatorial search space:
\begin{itemize}
    \item \textbf{Easy Level:} a small container (e.g., $2{\times}2{\times}3$) with relatively few pieces and limited shape complexity. Feasible placements are constrained and the solution space is small.
    \item \textbf{Middle Level:} a medium container (e.g., $2{\times}3{\times}3$) with increased spatial freedom and more placement permutations. The agent must reason more carefully about partial fills to avoid creating unreachable cavities.
    \item \textbf{Hard Level:} a large container (e.g., $3{\times}3{\times}4$) and a larger set of pieces. This setting introduces substantial ambiguity in placement order and orientation, making the task highly combinatorial and sensitive to early decisions.
\end{itemize}

\subsection{Case Analysis of Stacking Task}
\label{app:case_analysis}
\begin{figure}[htbp]
  \centering

  \begin{subfigure}[b]{0.23\textwidth}
    \centering
    \includegraphics[width=\linewidth]{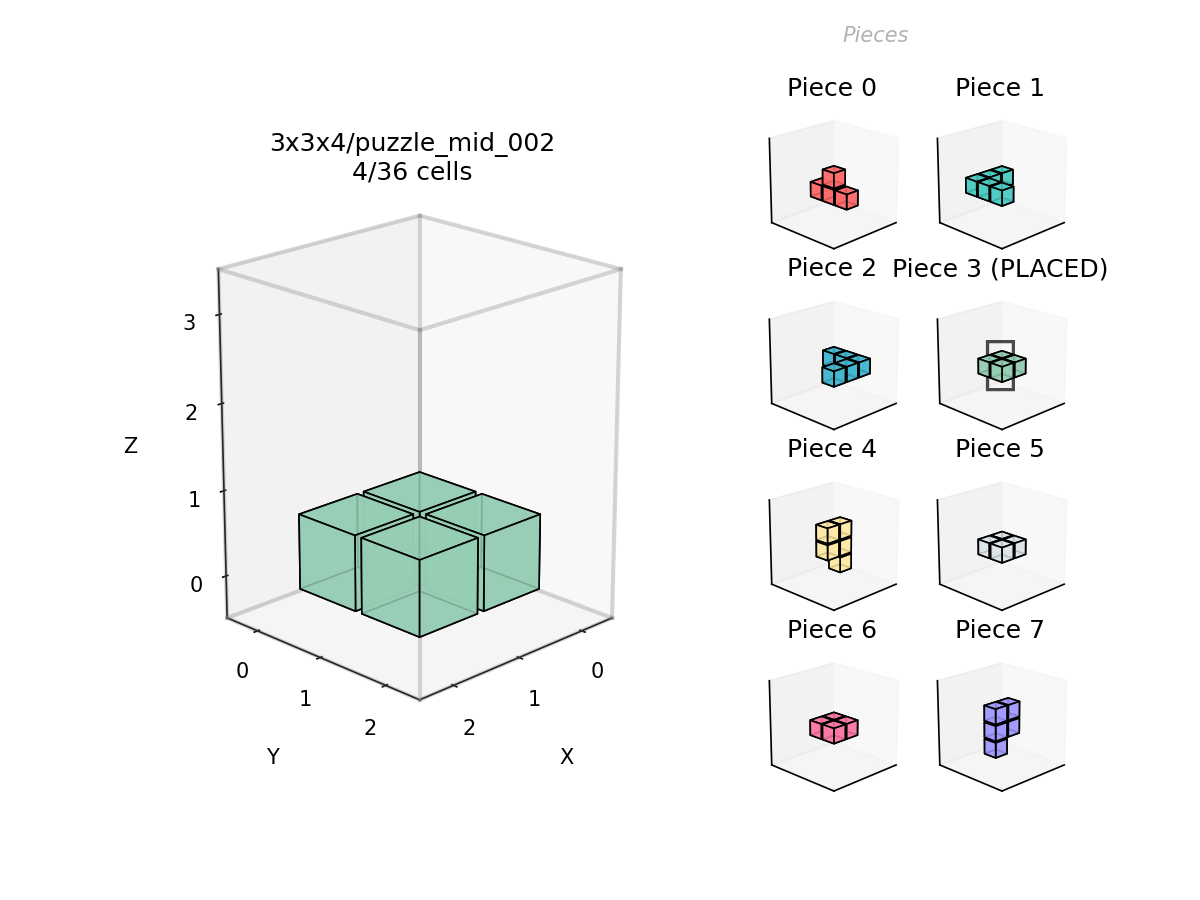}
    \caption{Step 1}
    \label{fig:sss1}
  \end{subfigure}\hfill
  \begin{subfigure}[b]{0.23\textwidth}
    \centering
    \includegraphics[width=\linewidth]{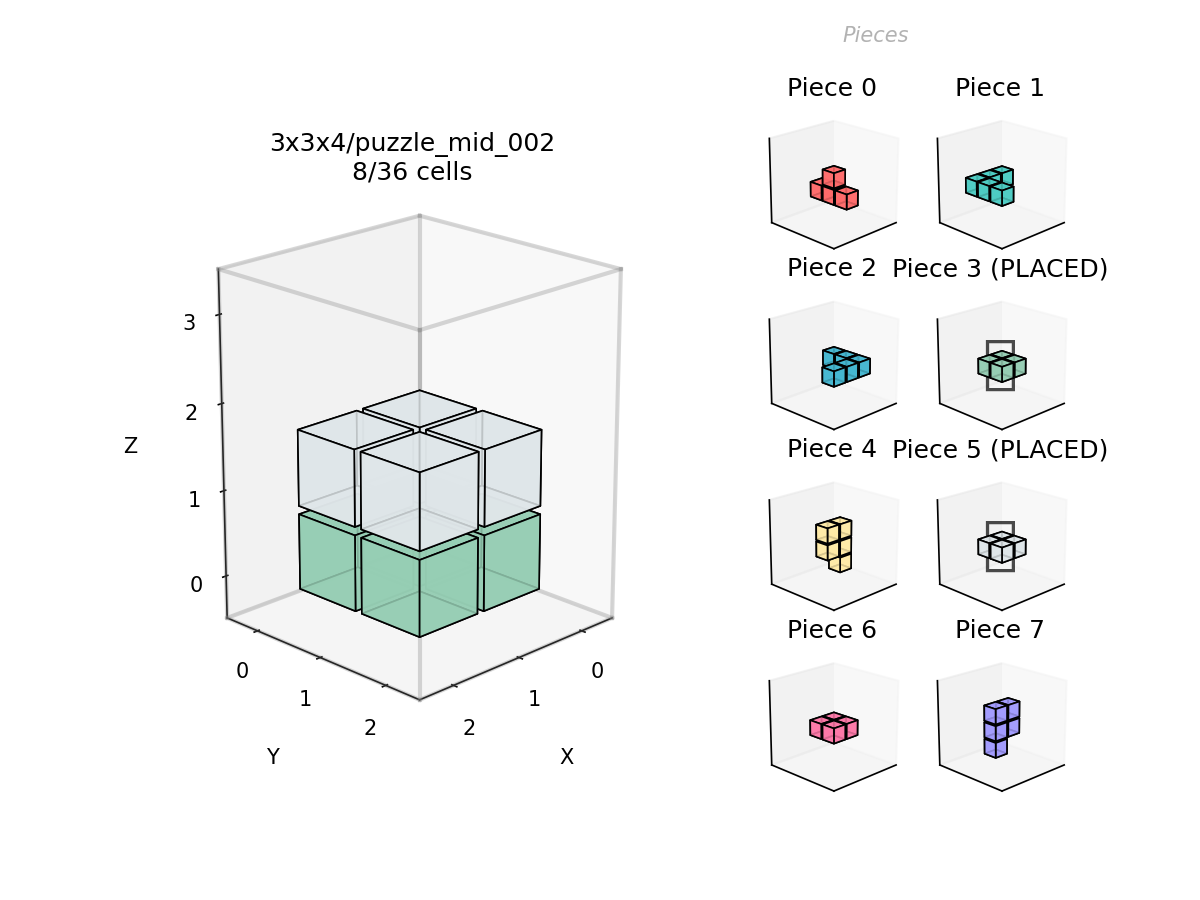}
    \caption{Step 2}
    \label{fig:sss2}
  \end{subfigure}\hfill
  \begin{subfigure}[b]{0.23\textwidth}
    \centering
    \includegraphics[width=\linewidth]{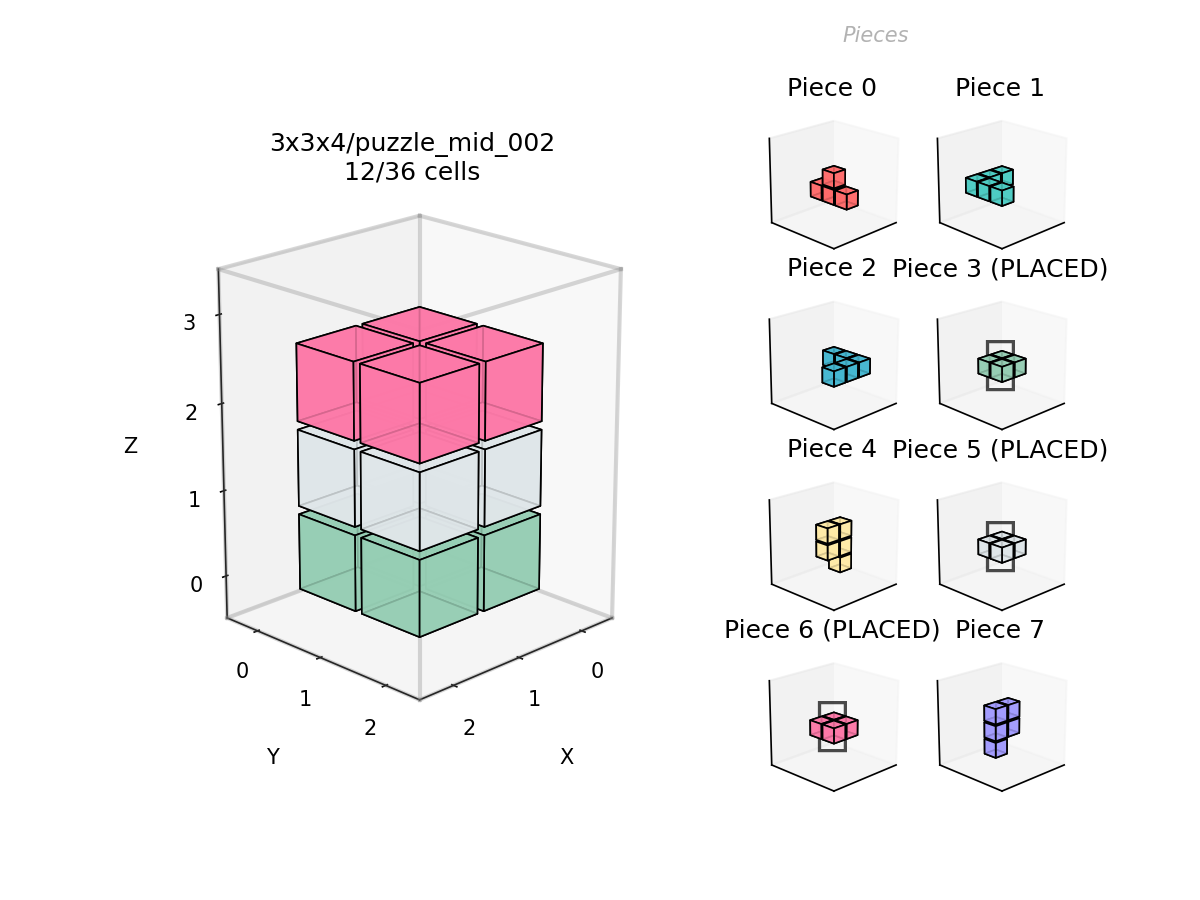}
    \caption{Step 3}
    \label{fig:sss3}
  \end{subfigure}\hfill
  \begin{subfigure}[b]{0.23\textwidth}
    \centering
    \includegraphics[width=\linewidth]{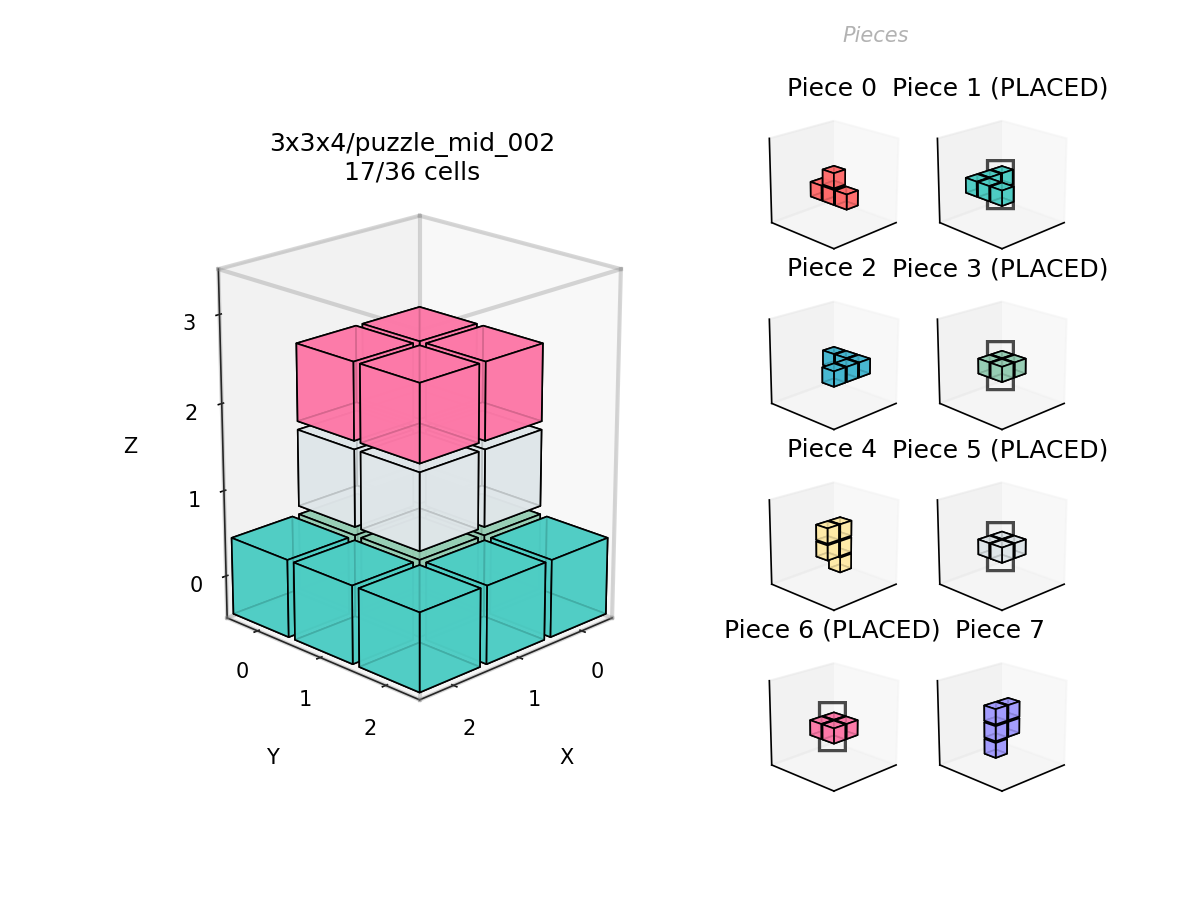}
    \caption{Step 4}
    \label{fig:sss4}
  \end{subfigure}

  \vspace{0.6em} 

  \begin{subfigure}[b]{0.23\textwidth}
    \centering
    \includegraphics[width=\linewidth]{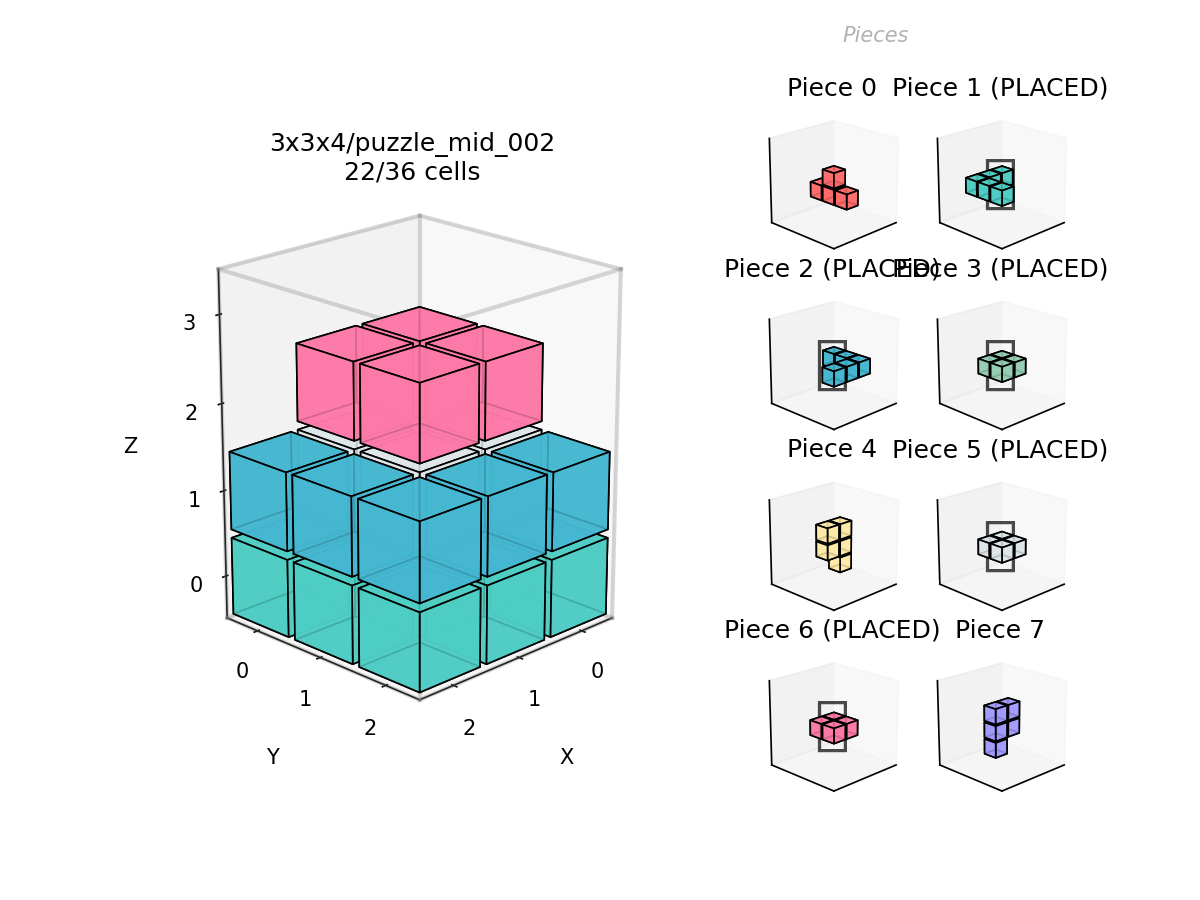}
    \caption{Step 5}
    \label{fig:sss5}
  \end{subfigure}\hfill
  \begin{subfigure}[b]{0.23\textwidth}
    \centering
    \includegraphics[width=\linewidth]{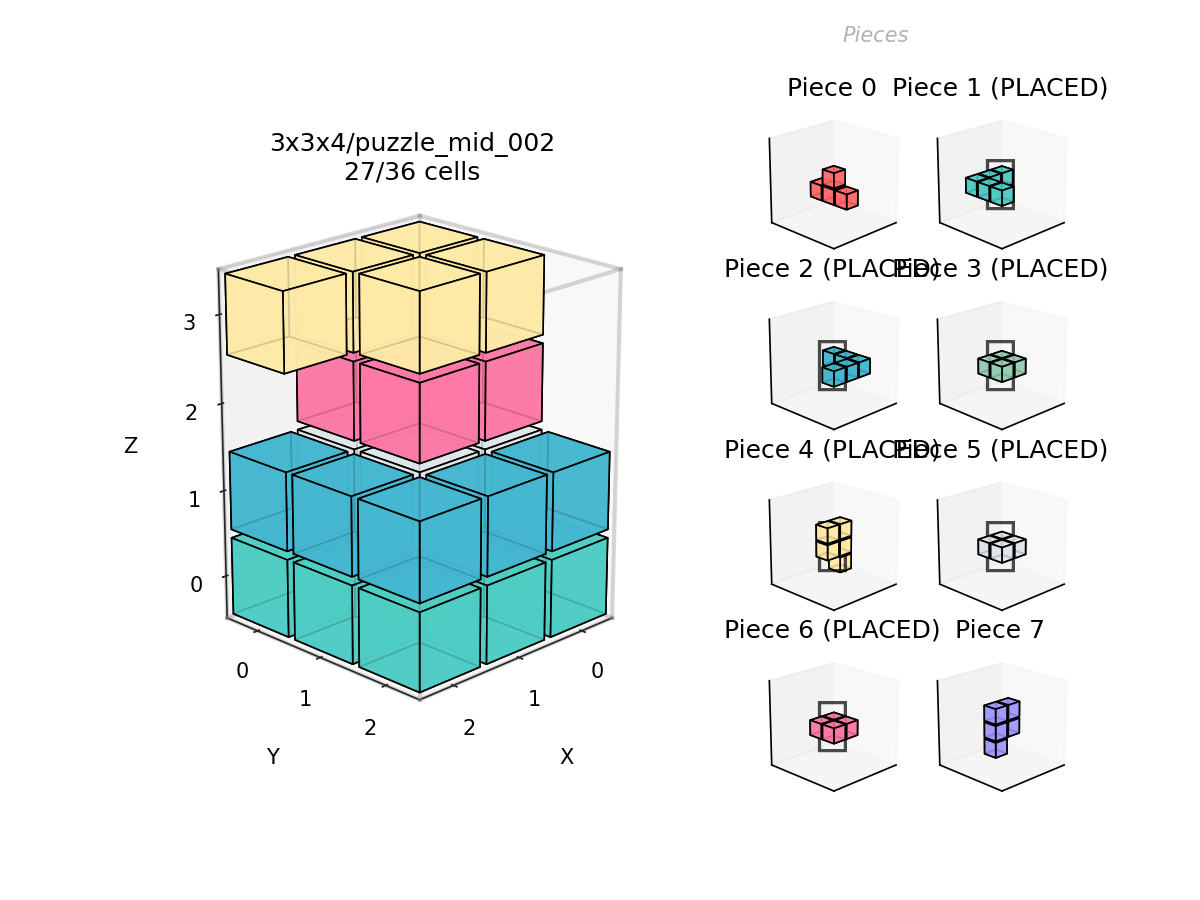}
    \caption{Step 6}
    \label{fig:sss6}
  \end{subfigure}\hfill
  \begin{subfigure}[b]{0.23\textwidth}
    \centering
    \includegraphics[width=\linewidth]{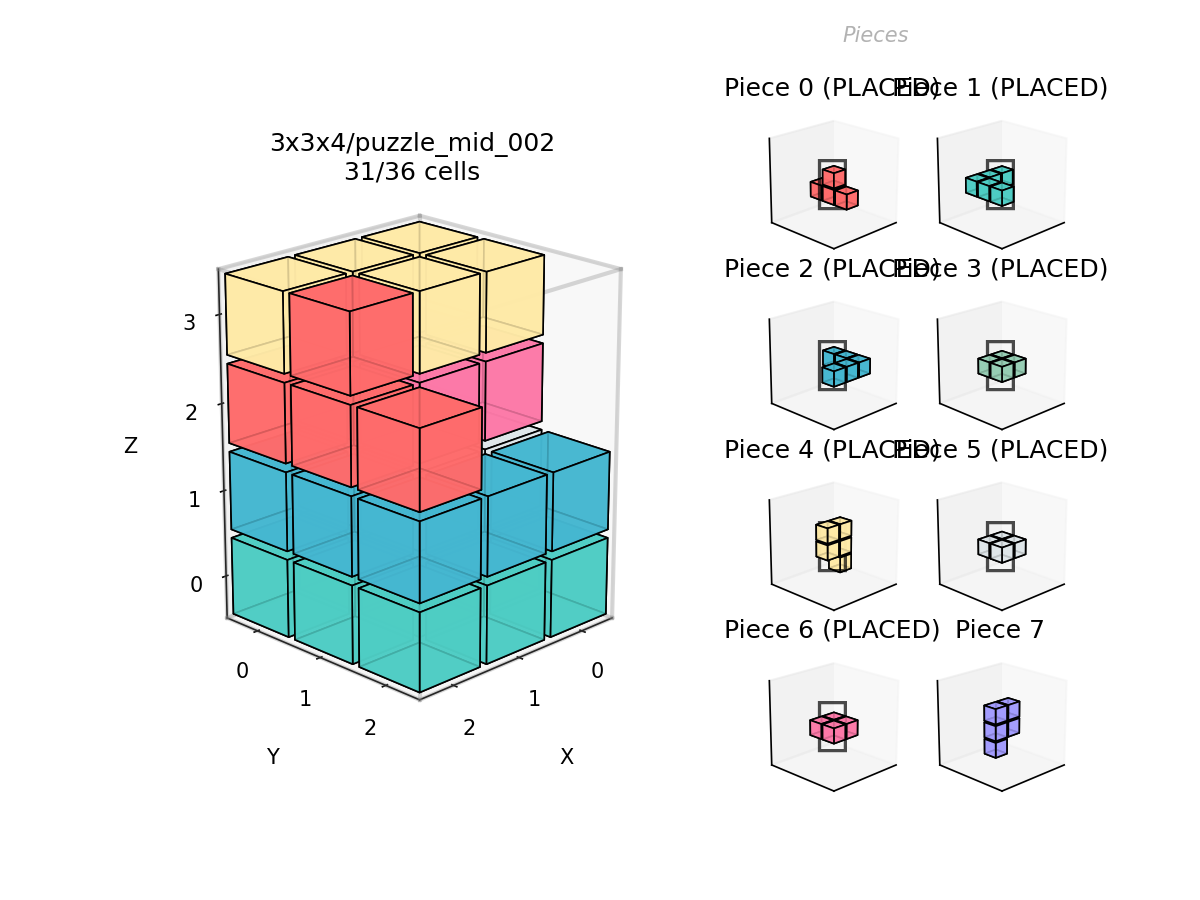}
    \caption{Step 7}
    \label{fig:sss7}
  \end{subfigure}\hfill
  \begin{subfigure}[b]{0.23\textwidth}
    \centering
    \includegraphics[width=\linewidth]{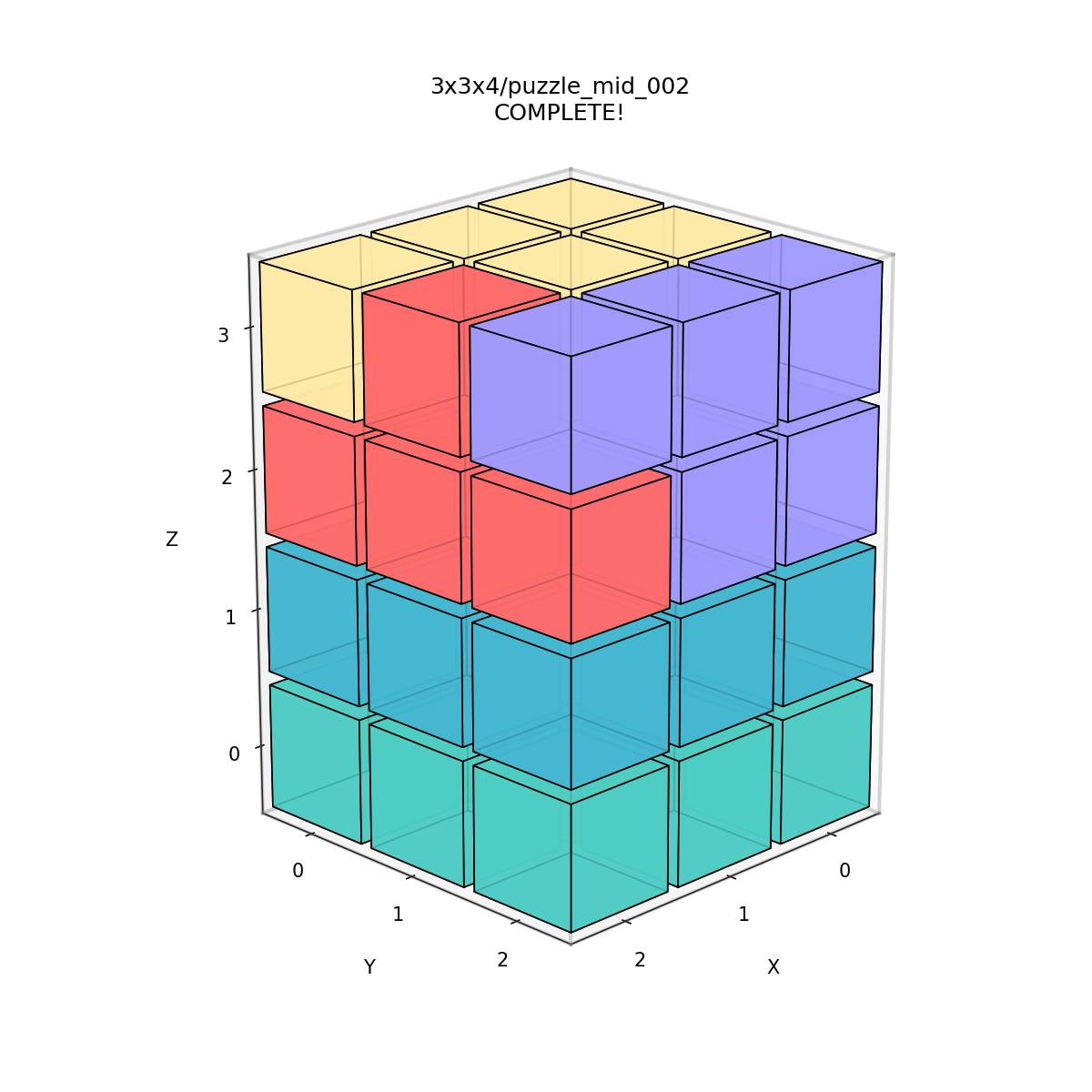}
    \caption{Step 8}
    \label{fig:sss8}
  \end{subfigure}

  \caption{A case of successfully solving the stacking task.}
  \label{fig:stacking_solved}
\end{figure}

\begin{figure}[htbp]
  \centering

  \begin{subfigure}[b]{0.23\textwidth}
    \centering
    \includegraphics[width=\linewidth]{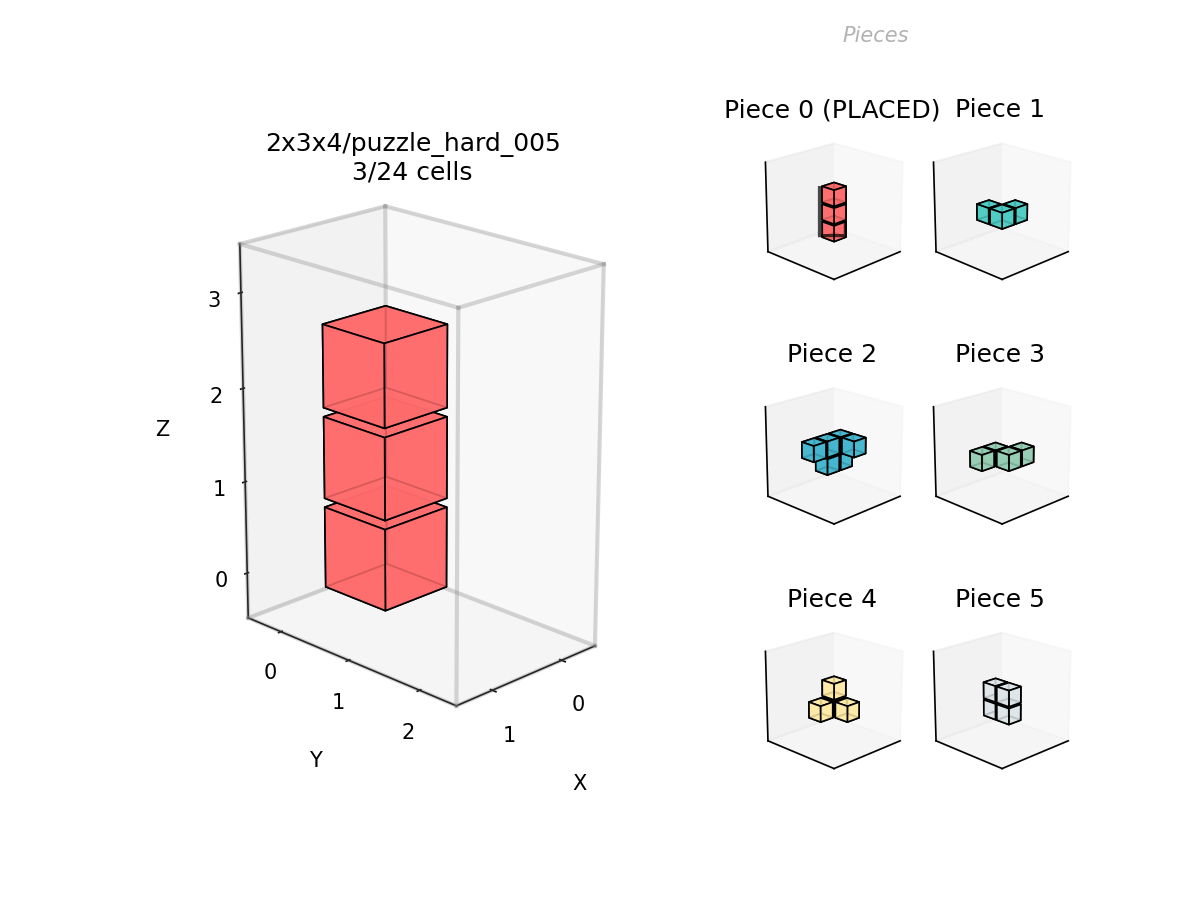}
    \caption{Step 1}
    \label{fig:sfs1}
  \end{subfigure}\hfill
  \begin{subfigure}[b]{0.23\textwidth}
    \centering
    \includegraphics[width=\linewidth]{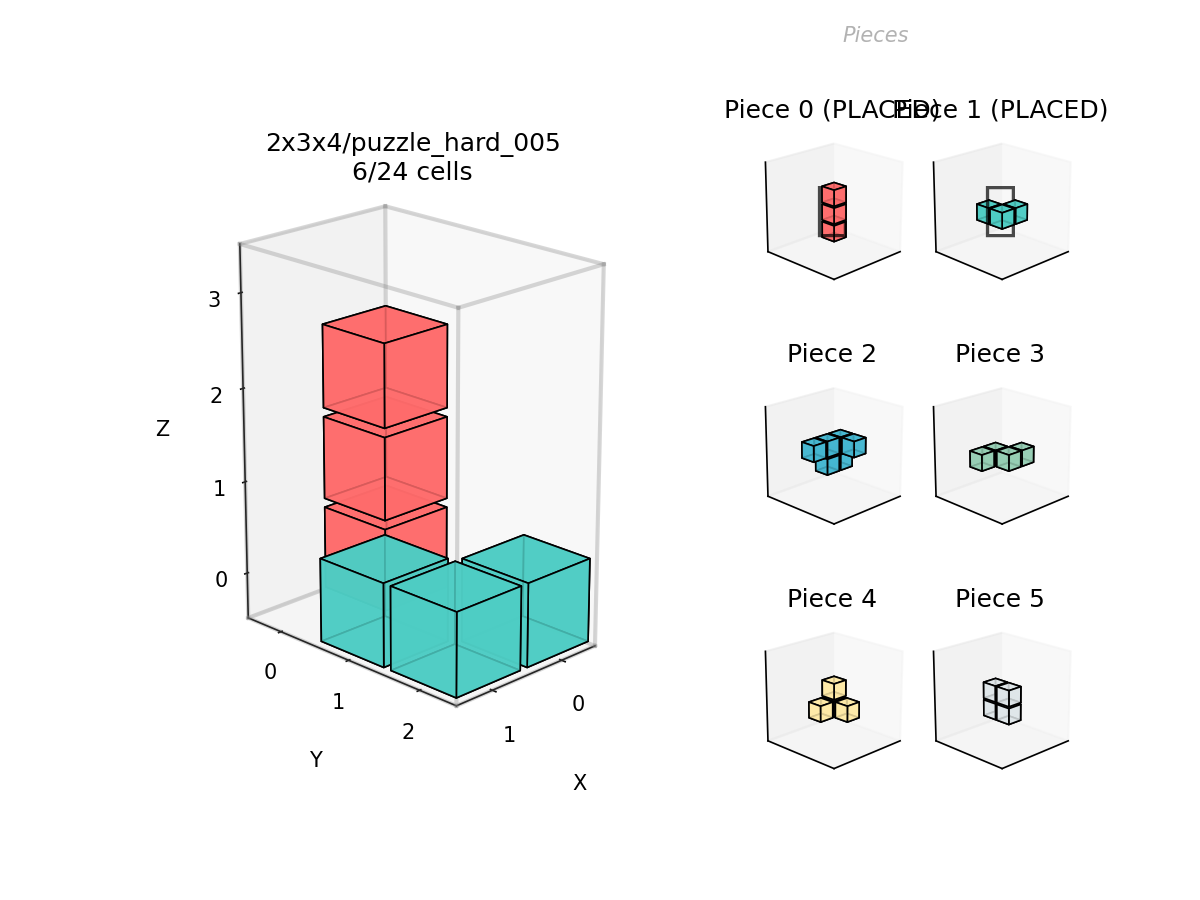}
    \caption{Step 2}
    \label{fig:sfs2}
  \end{subfigure}\hfill
  \begin{subfigure}[b]{0.23\textwidth}
    \centering
    \includegraphics[width=\linewidth]{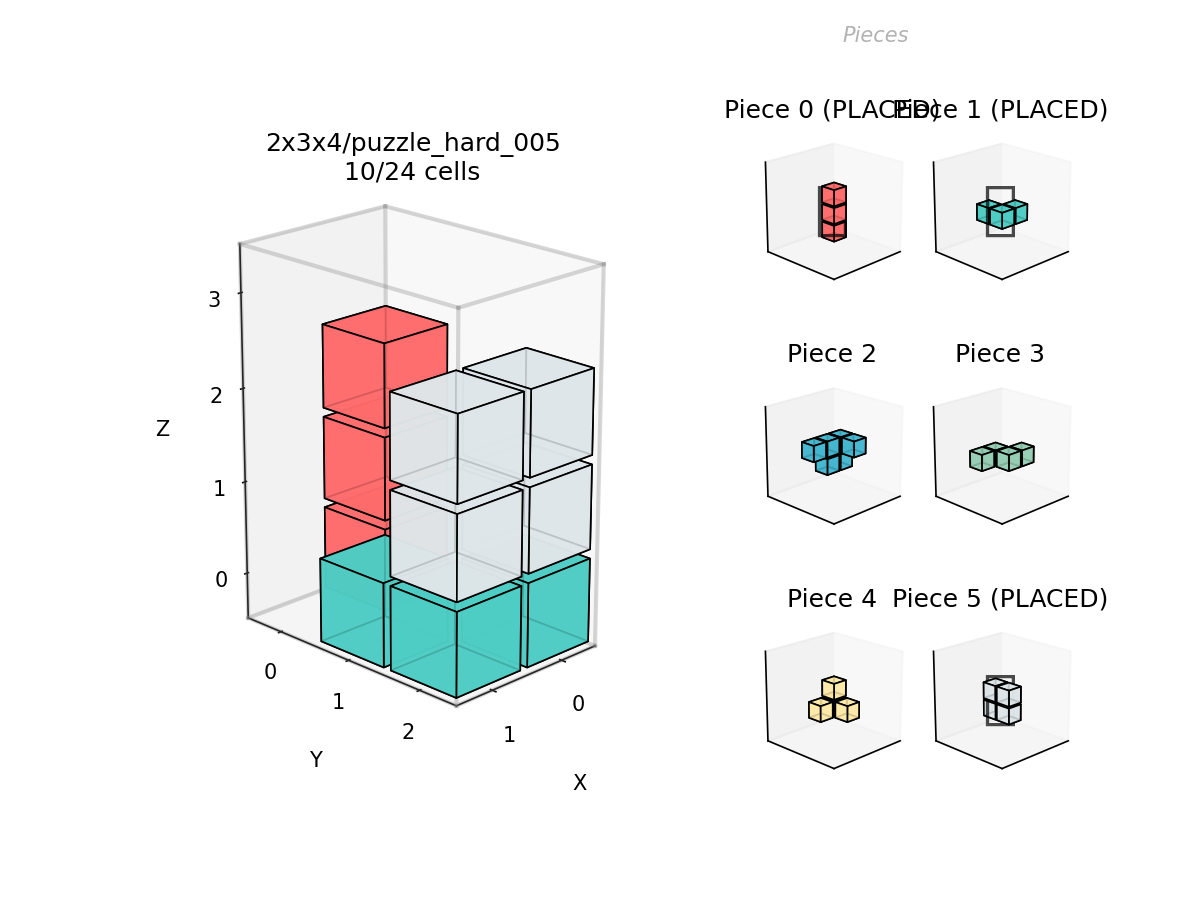}
    \caption{Step 3}
    \label{fig:sfs3}
  \end{subfigure}\hfill
  \begin{subfigure}[b]{0.23\textwidth}
    \centering
    \includegraphics[width=\linewidth]{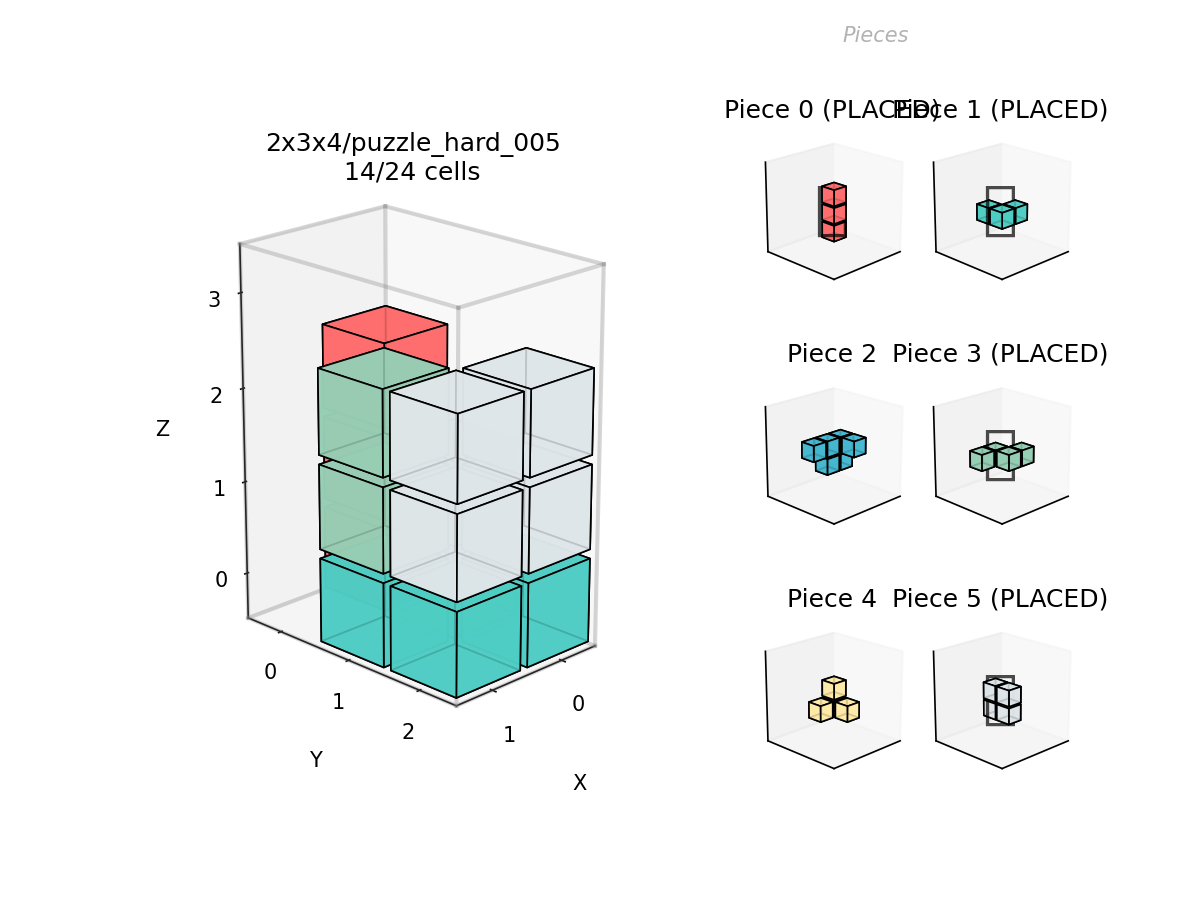}
    \caption{Step 4}
    \label{fig:sfs4}
  \end{subfigure}

  \vspace{0.6em} 

  \begin{subfigure}[b]{0.23\textwidth}
    \centering
    \includegraphics[width=\linewidth]{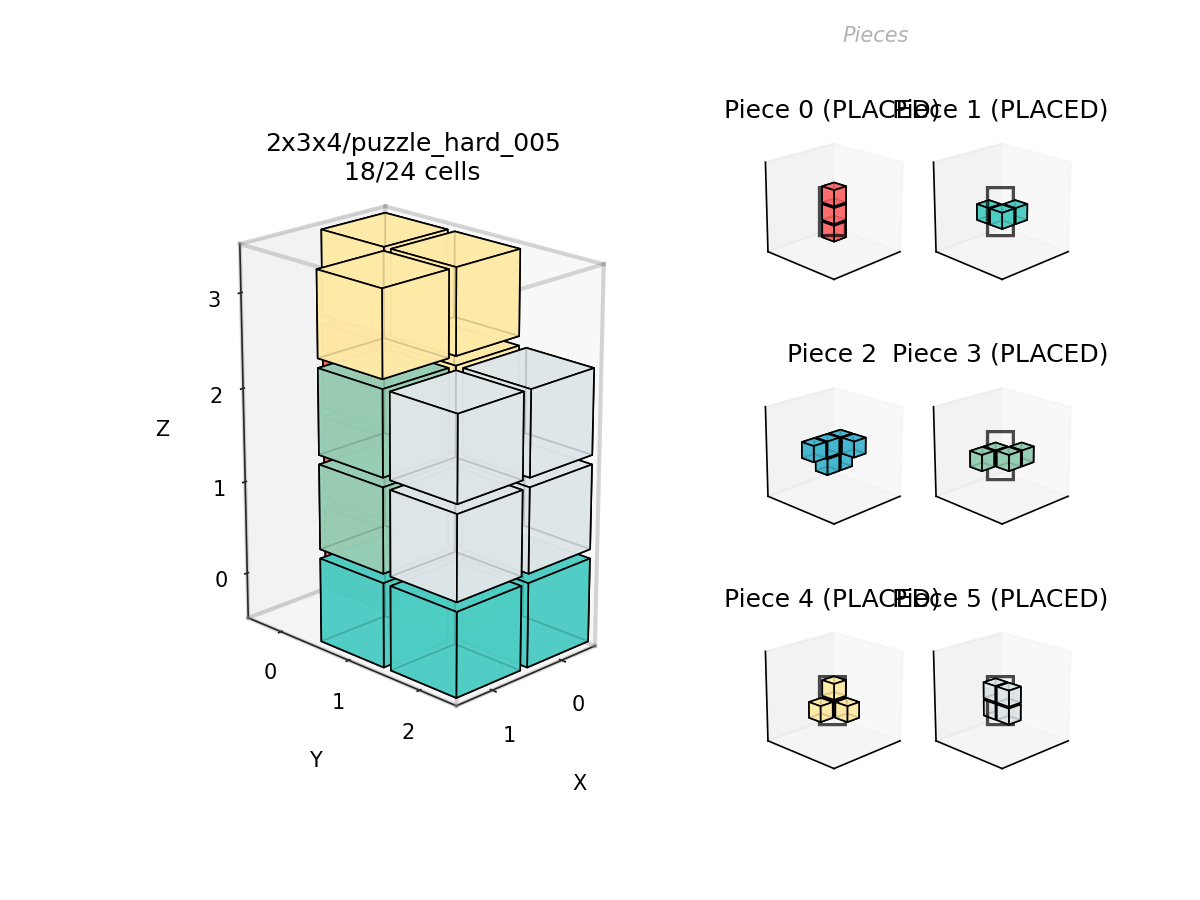}
    \caption{Step 5}
    \label{fig:sfs5}
  \end{subfigure}\hfill
  \begin{subfigure}[b]{0.23\textwidth}
    \centering
    \includegraphics[width=\linewidth]{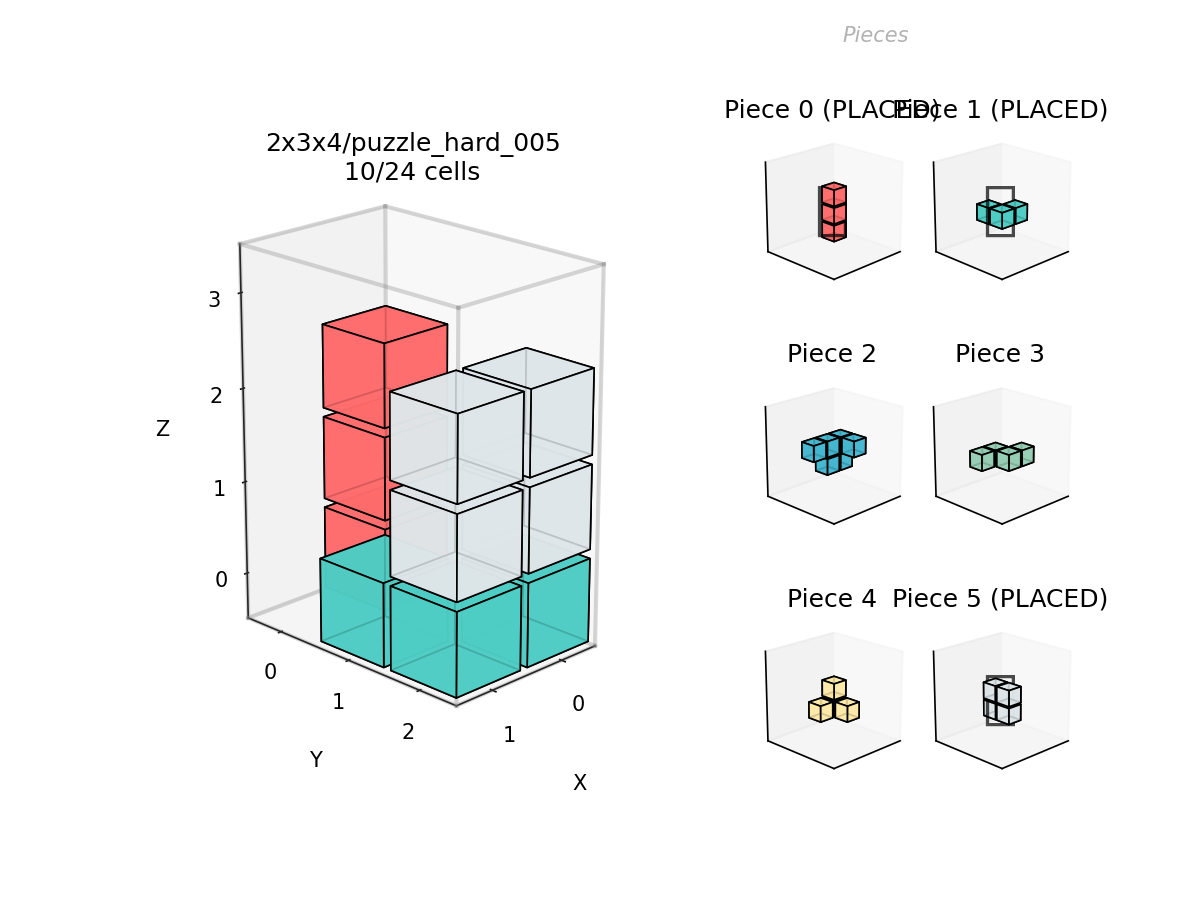}
    \caption{Step 6}
    \label{fig:sfs6}
  \end{subfigure}\hfill
  \begin{subfigure}[b]{0.23\textwidth}
    \centering
    \includegraphics[width=\linewidth]{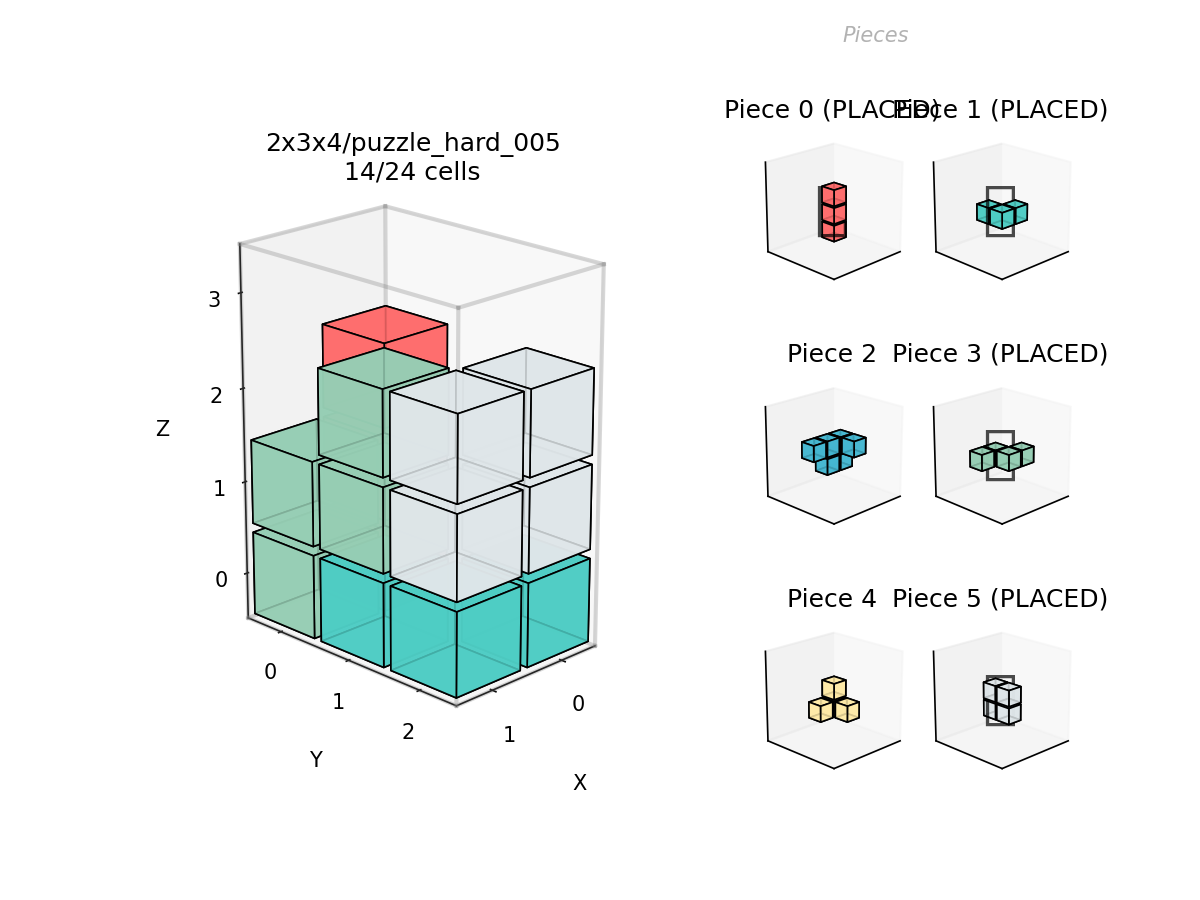}
    \caption{Step 7}
    \label{fig:sfs7}
  \end{subfigure}\hfill
  \begin{subfigure}[b]{0.23\textwidth}
    \centering
    \includegraphics[width=\linewidth]{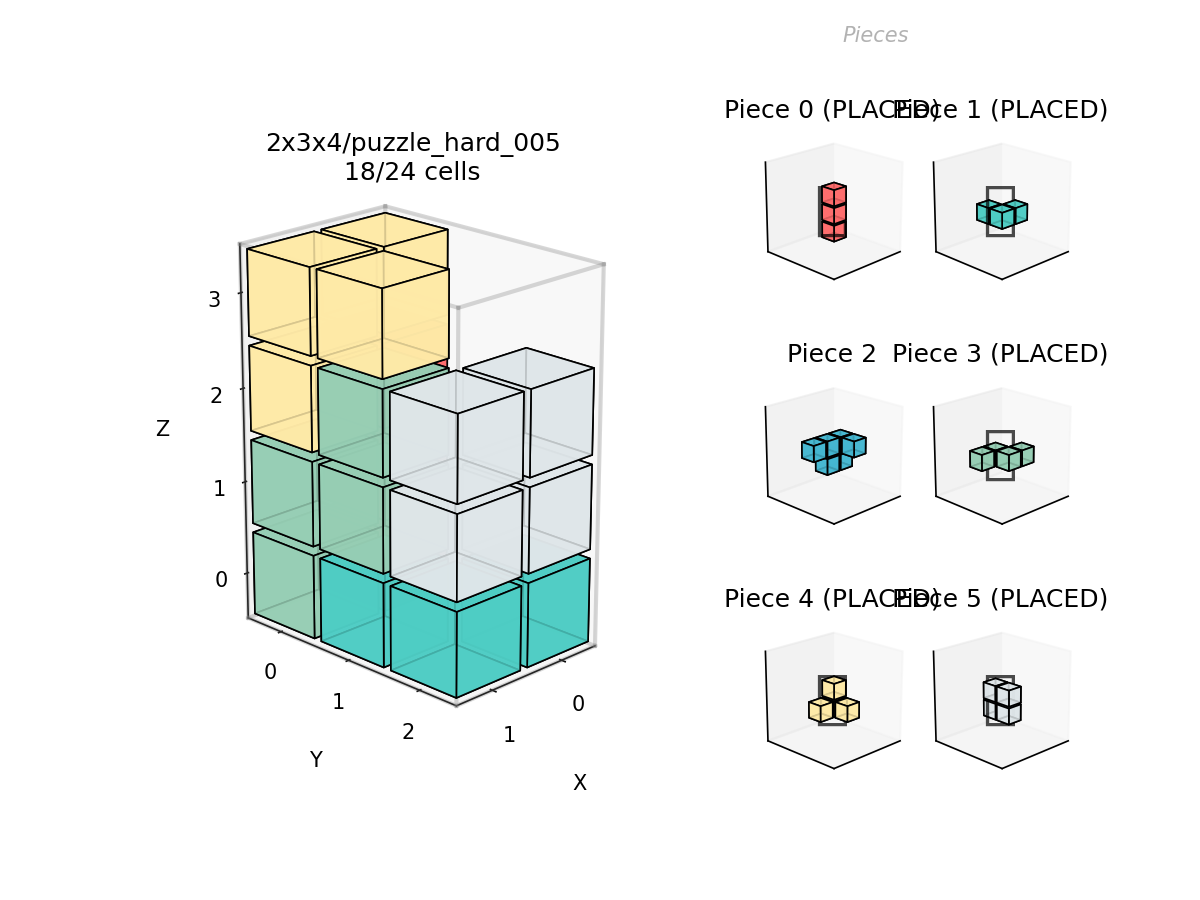}
    \caption{Step 8}
    \label{fig:sfs8}
  \end{subfigure}

  \caption{A case of failed to solve the stacking task.}
  \label{fig:stacking_failed}
\end{figure}

Fig.~\ref{fig:stacking_solved} and Fig.~\ref{fig:stacking_failed} demonstrate a successful and a failed trajectory of the stacking task, respectively. In the successful case (Fig.~\ref{fig:stacking_solved}), the agent follows a structure-first strategy. It starts by placing compact blocks to form a stable, flat base (Steps 1–2), and then stacks the upper layers layer by layer (Step 3). After the base is expanded (Step 4), subsequent placements (Steps 5–7) are used to systematically fill the remaining volume without creating unsupported overhangs or isolated cavities. This ordering preserves sufficient free space for later insertions and keeps the target shape “fillable” at every intermediate state, enabling the final completion.

In contrast, the failure case (Fig.~\ref{fig:stacking_failed}) is mainly caused by an early irreversible commitment. The agent places a tall vertical column at the beginning (Step 1), before establishing a complete foundation. This choice constrains where later pieces can go, forcing the following placements to wrap around the column (Steps 2–5). Even though the agent attempts to roll back some actions and try new placements (Steps 6–8), the partially built structure encloses awkward residual space and reduces the clearance needed to insert the remaining piece. Although each intermediate placement is locally valid, the resulting configuration becomes a dead-end: the remaining piece cannot fit due to geometric mismatch, so the episode ends incomplete.

\subsection{Polycube Stacking Puzzle Generation Pipeline}
\label{app:polycube-gen}

This appendix describes how we generate stable, solvable, and de-duplicated 3D polycube \emph{stacking/assembly} puzzles inside an $a \times b \times c$ voxel box. The generator follows a \emph{sample--verify} design: we first sample candidate partitions into connected pieces, then apply a sequence of hard validity checks (exact-cover solvability, structural constraints, and (optional) linear assembly feasibility). We also enforce strong de-duplication by canonicalizing each piece shape up to rigid rotations and hashing the resulting multiset.

\subsubsection{Objectives and Core Constraints}
\label{app:polycube-gen:objectives}

Given a box $\mathcal{B}=\{0,\dots,a\!-\!1\}\times\{0,\dots,b\!-\!1\}\times\{0,\dots,c\!-\!1\}$, we generate a set of pieces $\mathcal{P}=\{P_1,\dots,P_K\}$ such that:
\begin{enumerate}
  \item \textbf{Exact cover (solvable packing).} There exists a placement of all pieces (allowing 24 rigid rotations and translations within the box) that covers each voxel in $\mathcal{B}$ \emph{exactly once}.
  \item \textbf{Connectivity.} Each piece is 6-neighbor connected.
  \item \textbf{Piece size bounds.} Each piece satisfies $|P_i|\in[\texttt{min\_piece}, \texttt{max\_piece\_cells}]$.
  \item \textbf{Structural rule: no $2\times 3$ full face.} No piece contains a filled $2\times 3$ rectangle in any axis-aligned plane (XY/YZ/XZ), which avoids large flat plates that often reduce interaction richness.
  \item \textbf{No isolated piece.} In the final packed solution, each piece must touch at least one other piece via 6-neighbor adjacency (preventing ``floating'' blocks).
  \item \textbf{(Optional) Linear assembly.} If enabled, the packed solution must admit a collision-free \emph{linear} disassembly/assembly sequence where each step removes one piece by translating it along one of $\{\pm x,\pm y,\pm z\}$ without intersecting remaining pieces.
\end{enumerate}

\subsubsection{Candidate Partition Sampling by Difficulty}
\label{app:polycube-gen:sampling}

We support three difficulty modes by varying how candidate pieces are sampled before verification:

\paragraph{Easy (regular cuboids).}
We sample a multiset of axis-aligned cuboids whose volumes sum to $abc$ and whose dimensions satisfy an ``easy-shape'' rule: at least two dimensions are equal (e.g., $1\!\times\!1\!\times\!2$, $1\!\times\!2\!\times\!2$, $2\!\times\!2\!\times\!1$). We then verify whether these cuboids can be packed (with rotations) to exactly fill the box via the same exact-cover solver as other modes.

\paragraph{Mid (flat connected growth).}
We grow connected pieces by seeded BFS-style expansion, but constrain each piece to a plane by fixing one coordinate (e.g., $x=x_0$). This produces ``sheet-like'' structures while remaining connected and bounded in size.

\paragraph{Hard (free connected growth + uniqueness).}
We grow connected pieces without planar constraints and further enforce that all pieces have \emph{distinct} shapes up to rigid rotations. If duplicates occur, we attempt to improve uniqueness via limited voxel transfers between adjacent pieces while maintaining connectivity; otherwise we reject the sample. We also use solver search effort (visited nodes) as a difficulty proxy and enforce a mode-specific threshold.

\subsubsection{Exact-Cover Solvability via DLX}
\label{app:polycube-gen:dlx}

We validate packability by reducing the problem to an \emph{exact cover} instance. Each candidate placement of piece $i$ corresponds to a row that covers:
(i) the voxels occupied by the placement, and (ii) a ``use piece $i$'' column that enforces each piece is used exactly once.
We solve the resulting sparse exact-cover matrix using Algorithm~X with dancing links (DLX), augmented with:
(1) caching all placements per canonical shape and box size,
(2) an \emph{anchor piece} (the piece with fewest placements) constrained to cover the origin voxel to reduce symmetries,
and (3) an MRV-style column choice heuristic.

\subsubsection{Physical Feasibility and Linear Assembly Extraction}
\label{app:polycube-gen:assembly}

After DLX returns a packed solution (a concrete set of voxels for each placed piece), we apply two physical-feasibility checks.

\paragraph{No isolated piece.}
We reject solutions where any piece has no 6-neighbor contact with other pieces. This avoids degenerate instances that are technically solvable but physically uninformative for assembly reasoning.

\paragraph{Linear disassembly/assembly sequence.}
If linear assembly is required, we attempt to find a sequence in which pieces can be removed one-by-one by a pure translation along one of the six axis directions, without colliding with remaining pieces. A piece $P$ is \emph{removable} along direction $\mathbf{d}$ if translating each voxel of $P$ by one unit along $\mathbf{d}$ does not enter any voxel occupied by other pieces that are still present (within box bounds). We greedily remove any currently removable piece; if we can remove all pieces, reversing this removal order yields a valid assembly sequence.

Algorithm~\ref{alg:linear-assembly} gives the extraction procedure used in our pipeline.


\newcommand{\Removable}{\textsc{Removable}}

\begin{algorithm}[t]
\caption{Linear Assembly Sequence Extraction (via Greedy Disassembly)}
\label{alg:linear-assembly}
\begin{algorithmic}[1]
\Require Placed pieces $\{C_i\}_{i=1}^K$, where $C_i \subseteq \mathcal{B}$ is the set of voxels occupied by piece $i$
\Ensure Assembly order $\pi$ and removal directions $\{\mathbf{d}_t\}$, or \textsc{Fail}

\State $R \gets \{1,2,\dots,K\}$ \Comment{Indices of remaining pieces}
\State $\mathit{removal} \gets [\,]$ \Comment{Disassembly sequence as a list of $(i,\mathbf{d})$}

\While{$R \neq \emptyset$}
    \State $U \gets \bigcup_{j \in R} C_j$ \Comment{All voxels occupied by remaining pieces}
    \State $\mathit{found} \gets \textbf{false}$

    \For{$i \in R$}
        \State $O \gets U \setminus C_i$ \Comment{Voxels occupied by other pieces}
        \For{$\mathbf{d} \in \{+\vec{x},-\vec{x},+\vec{y},-\vec{y},+\vec{z},-\vec{z}\}$}
            \If{\Call{Removable}{$C_i, O, \mathbf{d}$}}
                \State Append $(i,\mathbf{d})$ to $\mathit{removal}$
                \State $R \gets R \setminus \{i\}$
                \State $\mathit{found} \gets \textbf{true}$
                \State \textbf{break}
            \EndIf
        \EndFor
        \If{$\mathit{found}$}
            \State \textbf{break}
        \EndIf
    \EndFor

    \If{\textbf{not} $\mathit{found}$}
        \State \Return \textsc{Fail}
    \EndIf
\EndWhile

\State $\pi \gets \Call{Reverse}{\mathit{removal}}$
\State \Return $\pi$
\end{algorithmic}
\end{algorithm}

\noindent
The predicate \textsc{Removable} checks whether translating $C_i$ by one voxel step along $\mathbf{d}$ causes any voxel of the translated set to intersect $O$ (within bounds). If no direction works for any remaining piece at some iteration, the instance does not admit a linear assembly and is rejected when linear assembly is required.

\subsubsection{Strong De-duplication by Shape Multiset Canonicalization}
\label{app:polycube-gen:dedup}

To avoid repeated puzzle templates, we de-duplicate instances by canonicalizing each piece shape up to rigid rotation.
For each piece $P_i$, we:
(i) translate it to a normalized origin (minimum coordinate at zero), then
(ii) enumerate its 24 rotation-equivalent shapes, and
(iii) select the lexicographically smallest voxel list as its canonical representation.
We then sort the list of canonical piece representations to form a \emph{multiset signature} for the puzzle and hash it (e.g., SHA1). In multi-processing, we reserve signatures with an atomic marker file to ensure thread-safe de-duplication.

\subsubsection{End-to-End Batch Generation and Outputs}
\label{app:polycube-gen:batch}

For each box size $(a,b,c)$ and difficulty mode (easy/mid/hard), we generate a fixed number of puzzles by repeatedly invoking the staged sampler with verification and de-duplication. Each accepted puzzle is stored in its own folder with:
(i) a JSON manifest (pieces, packed solution, assembly order, difficulty stats, signature), and
(ii) rendered 3D voxel visualizations of the full assembly and individual pieces.

Algorithm~\ref{alg:batch-gen} summarizes the full pipeline.

\begin{algorithm}[t]
    \caption{Batch Generation with Staged Sampling, Verification and De-duplication}
    \label{alg:batch-gen}
    \begin{algorithmic}[1]  
        \Require
            Set of box sizes $\mathcal{D}$, \newline  
            Per-size configuration (max pieces, difficulty thresholds), \newline
            Target number of puzzles per size/mode $T$
        \Ensure
            A dataset of accepted puzzles, \newline
            Summary metadata (counts, rejection reasons, generation statistics)

        \ForAll{$(a,b,c)\in \mathcal{D}$}
            \For{each mode $\in \{\texttt{easy}, \texttt{mid}, \texttt{hard}\}$}
                \For{$t = 1$ to $T$}
                    \State $\textit{gen} \gets$ \Call{GenerateOneStaged}{$a, b, c, \textit{mode}, \textit{cfg}$}
                    \If{$\textit{gen} = \texttt{Fail}$}
                        \State \textbf{continue}
                    \EndIf

                    \State $s \gets$ \Call{PuzzleSignature}{$\textit{gen}.\textit{pieces}$}
                    \If{$\neg$ \Call{ReserveSignatureAtomically}{$s$}}  
                        \State \textbf{continue}
                    \EndIf

                    \State \Call{SavePuzzle}{$\textit{gen}, s$}
                    \Comment{Store as JSON + visual renders in per-puzzle folder}
                \EndFor
            \EndFor
        \EndFor

       \State \Return $(\mathit{dataset}, \mathit{metadata})$

    \end{algorithmic}
\end{algorithm}

\paragraph{Staged min-piece fallback.}
GenerateOneStaged implements a staged attempt schedule: it first searches with a stricter minimum piece size (e.g., $\texttt{min\_piece}=4$) and, if unsuccessful within budget, relaxes to $\texttt{min\_piece}=3$ to improve yield for small volumes or tight constraints. Each sampled partition is accepted only if it passes the full chain of checks:
\emph{(i) structural rules, (ii) DLX solvability, (iii) no isolated piece, (iv) (optional) linear assembly, and (v) difficulty threshold (hard).}

\subsection{Prompt details}
\label{app:world_model_prompt}
Fig. \ref{fig:world_model_prompt} shows the content of our world model prompt.


\begin{figure}[t]
    \centering
    \captionsetup{
        width=0.95\linewidth,
        font=small,
        skip=6pt,
        justification=justified
    }

    \caption{World-model prompt for generating physically valid disassembly videos of Kongming/Luban locks (interlocking burr puzzles).
    Constraints include rigidity, collision avoidance, continuous motion, and axis-aligned sliding.}
    \label{fig:world_model_prompt}

    \begin{tcolorbox}[
        width=0.95\linewidth,
        colback=white,
        colframe=gray!70!black,
        arc=3pt,
        boxrule=0.5pt,
        left=8pt, right=8pt,
        top=6pt, bottom=6pt
    ]
    \ttfamily\small
Generate a video showing the disassembly of a Kongming/Luban lock (interlocking burr puzzle)
from the \emph{exact} configuration in the provided reference image.

The video must be physically valid and obey these constraints:
\begin{itemize}[leftmargin=12pt, itemsep=2pt, topsep=2pt]
    \item All parts are rigid wooden blocks (no bending or deformation).
    \item No interpenetration: contacts and collisions are respected; parts never pass through each other.
    \item No teleportation: motion is continuous over time.
    \item Each piece moves only by axis-aligned sliding along its allowed rail; no rotation until it is fully freed.
    \item Use a realistic unlocking sequence: remove the key piece first, then remove any newly unblocked pieces until fully separated.
    \item Match the reference geometry and part count exactly; do not add, remove, resize, or alter pieces.
\end{itemize}
    \end{tcolorbox}
\end{figure}


\end{document}